\documentclass{macro/neu_msthesis}

\usepackage{todonotes}
\usepackage{caption}
\usepackage{subcaption}

\usepackage[doi=false, url=false, isbn=false, eprint=false]{biblatex}[IEEEtran]
\title{Progress Towards Untethered Autonomous Flight of Northeastern University's Aerobat}

\author{Adarsh Salagame}

\nuid{001096009}

\dept{Electrical and Computer Engineering}

\degreename{Robotics}

\field{Electrical and Computer Engineering}

\submitdate{August 2022}

\numberofmembers{3}
\principaladviser{Dr. Alireza Ramezani}
\firstreader{Dr. Hanumanth Singh}
\secondreader{Dr. Lawson Wong}
\thirdreader{Dr. Third Member}
\fourthreader{Dr. Fourth Member}
\fifthreader{Dr. Fifth Member}
\chairman{Dr. Srinivas Tadigadapa}
\dean{Dr. Waleed Meleis}

\usepackage{amsfonts,amssymb,amsmath}			
\usepackage{times}		
\usepackage{bm}

\newcommand{\ifno}[1]{}

\usepackage{multirow}
\makeindex
\usepackage{makeidx}
\usepackage{acronym}

\usepackage{url}
\ifnum\pdfoutput>0
\usepackage{epstopdf}
\else
\usepackage[pdftex]{graphicx}
\fi

\ifnum\pdfoutput>0
\usepackage[pdftex,
bookmarks=true,
bookmarksnumbered=true,
hypertexnames=false,
breaklinks=true           
]{hyperref}
\else
\usepackage[hypertex,
bookmarks=true,
bookmarksnumbered=true,
hypertexnames=false,
breaklinks=true           
]{hyperref}
\fi

\hypersetup{				
pdfauthor = {\authorRef},
pdftitle = {\titleRef},
pdfsubject = {\expandafter{\degreeRef} thesis submitted to Northeastern University},
pdfkeywords = {add keywords here}
pdfcreator = {LaTeX with hyperref package},
pdfproducer = {dvips + ps2pdf}}

\usepackage{microtype}

\begin{document}

\pdfbookmark[1]{Cover}{cover}

\titlepage

\begin{frontmatter}


\begin{dedication}
To my family.
\end{dedication}


\pdfbookmark[1]{Table of Contents}{contents}
\tableofcontents
\listoffigures
\newpage\ssp
\listoftables


\chapter*{List of Acronyms}
\addcontentsline{toc}{chapter}{List of Acronyms}

\begin{acronym}
\acro{FWMAV}{Flapping Wing Micro Aerial Vehicle}
\acro{BLDC Motor}{Brushless DC Motor}
\acro{ESC}{Electronic Speed Controller}
    Used to control the three-phase voltage to a BLDC motor to control its speed.
\acro{IMU}{Inertial Measurement Unit}
\acro{VIO}{Visual Inertial Odometry}
\acro{PWM}{Pulse Width Modulation}
\acro{FPV}{First Person View}
\acro{GPU}{Graphics Processing Unit}
\acro{ROS}{Robotic Operating System}
\end{acronym}


\begin{acknowledgements}

This work has been supported by a lot of people in a number of different ways. I would first like to acknowledge the mentorship and guidance provided to me by Dr. Alireza Ramezani over the course of this thesis, opening up new opportunities and perspectives for research. I would also like to acknowledge Dr. Hanumanth Singh and Dr. Milad Ramezani for their guidance. As a collaborative project involving multiple specialities, a number of students have contributed to make this work possible. I would like to acknowledge the foundational work done by Eric Sihite at Caltech University, and Xintao Hu and Bozhen Li at Northeastern University. I would also like to acknowledge the work done by Roman Snegatch, Hamza Iqbal, Arunbhaarthi Anbu, Rohit Rajput, Yizhe Xu and Xuejian Niu, and the brainstorming and troubleshooting support from the rest of the Silicon Synapse Lab. Finally, I would like to acknowledge my parents and everyone who made this work possible through their constant motivation and heartfelt support. 
\end{acknowledgements}


\begin{abstract}
State estimation and control is a well-studied problem in conventional aerial vehicles such as multi-rotors. But multi-rotors, while versatile, are not suitable for all applications. Due to turbulent airflow from ground effects, multi-rotors cannot fly in confined spaces. Flapping wing micro aerial vehicles have gained research interest in recent years due to their lightweight structure and ability to fly in tight spaces. Further, their soft deformable wings also make them relatively safer to fly around humans. This thesis will describe the progress made towards developing state estimation and controls on Northeastern University's Aerobat, a bio-inspired flapping wing micro aerial vehicle, with the goal of achieving untethered autonomous flight. Aerobat has a total weight of about 40g and an additional payload capacity of 40g, precluding the use of large processors or heavy sensors. With limited computation resources, this report discusses the challenges in achieving perception on such a platform and the steps taken towards untethered autonomous flight.

\end{abstract}

\end{frontmatter}

\pagestyle{headings}

\chapter{Introduction}
\label{chap:introduction}

Flapping Wing aerial locomotion is an interesting field of study that is gaining a lot of research interest \cite{di_luca_bioinspired_2017, eguiluz_towards_2019, phan_kubeetle-s_2019, chukewad_robofly_2021}. Flapping robots offer a number of advantages over conventional aerial robots such as quad-copters, which rely on propeller based lift generation. The biggest of these is their ability to fly in confined spaces. Quad-copters and other multi-rotor vehicles are heavily affected by turbulent air flow when flying in confined spaces or close to the ground \cite{matus-vargas_ground_2021}. On the other hand, flapping wing robots have the opposite effect, not only being able to fly in tight spaces aided by their high agility, but also showing higher efficiency when flying close to the ground, a phenomenon well studied in birds \cite{rayner_aerodynamics_1991}. This makes flapping wing robots a huge potential asset for applications in disaster management, for example flying through the narrow spaces inside a collapsed building, for applications in inspection such as flying through sewers or air vents that are inaccessible to humans and other types of robots, or even for data collection for scientific research in previously inaccessible areas. A further advantage of flapping wing robots is their relative safety to operate. With soft deformable wings and significantly smaller weight density, they are not only safer than propeller based aerial robots to operate around people, they are less affected by crashes into walls or ceilings and can continue flying. And finally, flapping wing robots are extremely agile, able to perform zero momentum turns, and are more efficient in their agility when compared with multi-rotor systems that rely on thrust vectoring for their agility, which is very power hungry. \cite{de_croon_flapping_2020, tu_acting_2019}

For all these advantages, however, flapping wing robots still pose a number of challenges that must be solved before they may fully reach the impact that multi-rotors have had. Flapping wing systems generate much less thrust when compared to multi-rotors of similar size. This severely impacts the available payload for sensors and other electronics that would enable the robot to be fully autonomous. Further, these are highly dynamic platforms, with flapping motions causing vibrations that an onboard perception system must deal with \cite{eguiluz_towards_2019}. Also, unlike multi-rotors, flapping systems have a constantly shifting center of mass, affected not only by the wing position, but also by the variable deformations in the wings and any inherent compliance in their structure due to their lightweight designs. These factors make localization and autonomous control of the robot a challenge.

In order to develop autonomous flight, two things are required:
\begin{enumerate}
    \item \textbf{Low Level Control}: The ability to track any desired trajectory and accurately execute any desired motion
    \item \textbf{High level control}: The ability to decide what trajectory or motion to execute based on knowledge about the robot state and it's surroundings. High level control may be further divided into two sub-goals:
    \begin{enumerate}
        \item \textbf{Perception and State Estimation}: Understand the surrounding environment and localize the robot within this space
        \item \textbf{Trajectory Planning}: Decide a trajectory to follow based on the perception and state estimation
    \end{enumerate}
\end{enumerate}

All of these are eventual goals for Aerobat. However, this work focuses on making progress towards Perception, State Estimation and Low-level control.

The thesis is organized according to these goals as follows. Chapter \ref{chap:related} goes through contemporary works on aerial and flapping wing systems, focusing specifically on works that have had success with autonomous flight. Chapter \ref{chap:untethered} describes initial results with open loop untethered flight and the development made towards safe and controlled testing of untethered flight. Chapter \ref{chap:control} describes the progress made towards low level control of Aerobat, describing the aerodynamic model of Aerobat and validation of the aerodynamic model. Chapter \ref{chap:perception} describes the progress made towards developing onboard perception and state estimation, with a special focus on the limited payload capacity available and the challenges in implementation on limited computation hardware. Finally, Chapter \ref{chap:conclude} presents an overview of the milestones reached, challenges faced and future development to take place towards untethered autonomous flight.

\section{About Aerobat}

Northeastern University's Aerobat is a tail-less flapping wing robot that, unlike existing examples, is capable of significantly morphing it's wing structure during each gait cycle. The robot, with a weight of 40g (when carrying a battery and a basic microcontroller) and a wingspan of 30 cm, was initially developed to study the flapping-wing flight of bats.

Aerobat utilizes a computational structure, called the \textit{Kinetic Sculpture} (KS) \cite{sihite_computational_2020}, that introduces computational resources for wing morphing. The KS is designed to actuate the robot's wings as it is split into two wing segments: the proximal and distal wings, which are actuated by what is the equivalent of shoulder and elbow joints, respectively. The gait captures the wing folding during the upstroke motion, which is one of the key modes in bat flight. The wing folding reduces the wing surface area and minimizes the negative lift during the upstroke and results in a more efficient flight. Aerobat is capable of flapping at a frequency of up to 8 Hz. Without a tail, Aerobat is unstable in its longitudinal (pitch dynamics) and frontal (roll dynamics) planes of flight. 

\chapter{Related Work}
\label{chap:related}

\begin{figure*}[t]
    \centering
    \includegraphics[width=\linewidth]{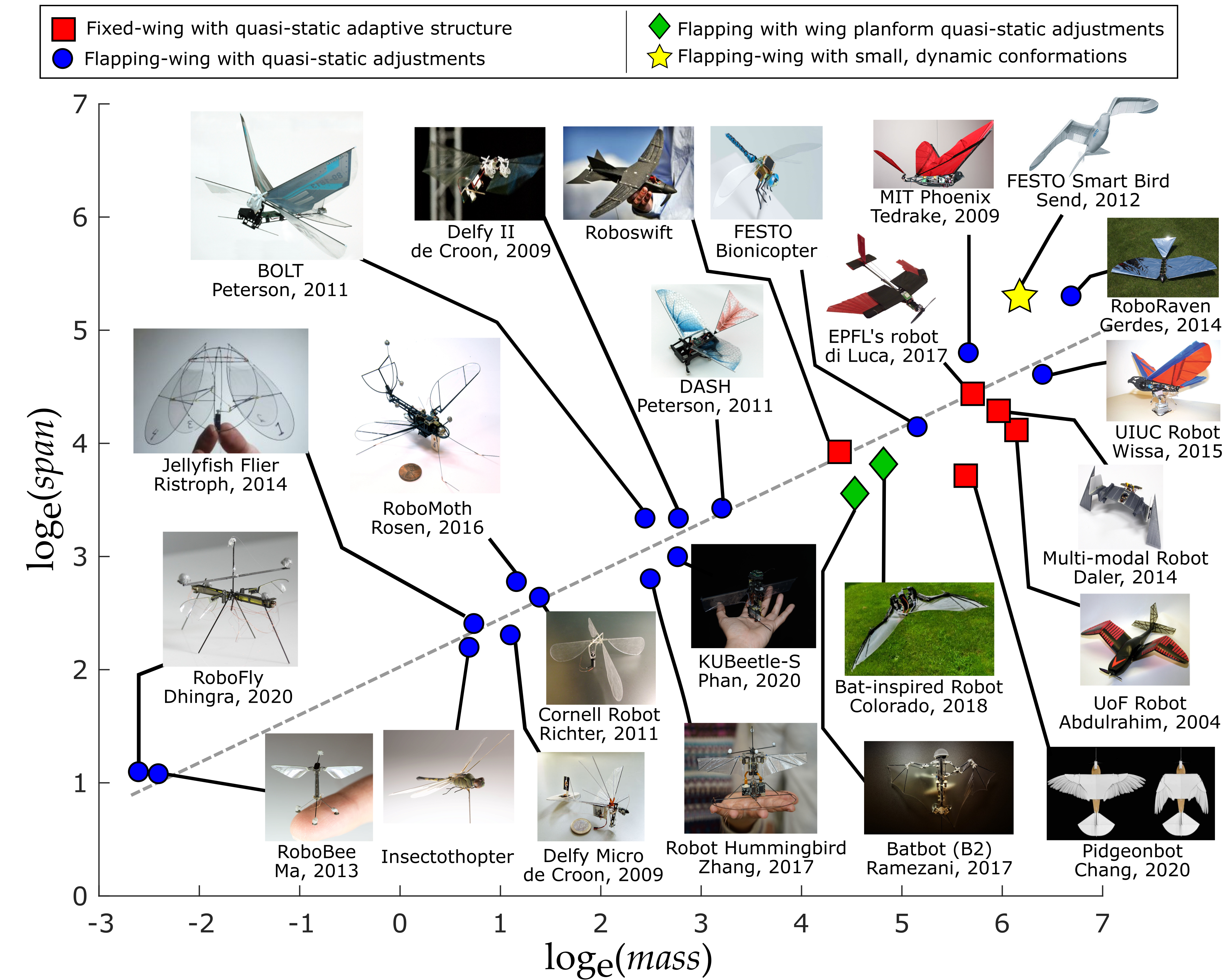}
    \caption{Illustrates state-of-the-art Micro Aerial Vehicle (MAV) designs classified based on wing morphing capabilities.}
    \label{fig:literature_search}
\end{figure*}

Flapping Wing Micro-Aerial Vehicle (FWMAV) platforms in the literature may be broadly divided into two categories based on the size of the robot. Insect-scale platforms such as Harvard's Robobee \cite{ma_controlled_2013} and University of Washington's RoboFly \cite{chukewad_robofly_2021} range in weight from a few milligrams to a few grams ($<$10g). These typically implement offboard processing with little to no payload budget for onboard sensors. Other examples of platforms in this category are Robo Moth \cite{rosen_development_2016}, Delfly Micro \cite{de_croon_design_2009}, Jellyfish Flier \cite{ristroph_stable_2014} and Insectothopter \cite{noauthor_insectothopter_nodate}.

Larger-scale platforms such as TU Delft's DelFly \cite{de_wagter_autonomous_2014} and Purdue's Hummingbird \cite{zhang_design_2017} weigh in the order of tens of grams and are capable of carrying sensors and sufficient processing onboard for basic perception. Other platforms in this category are UC Berkeley's DASH and BOLT \cite{peterson_experimental_2011} and KUBeetle-S \cite{phan_kubeetle-s_2019}. 

On the extreme end of this scale are large ornithopters such as the University of Seville's GRIFFIN \cite{zufferey_design_2021}, RoboRaven \cite{gerdes_robo_2014}, FESTO Smart Bird, Pidgeonbot \cite{chang_soft_2020}, MIT Phoenix \cite{tedrake_learning_nodate} and EPFL's morphing wing robot \cite{di_luca_bioinspired_2017} which all weigh in the order of a few hundred grams, comparable in weight and payload capacity to multi-rotors. All these platforms are compared in Figure \ref{fig:literature_search}by plotting them on a logarithmic scale of mass and wingspan.

Northeastern University's Aerobat sits uniquely in the middle of the range of FWMAV sizes. With a wing span of 30 cm and weighing 40g (when carrying a battery and a basic microcontroller), it is small enough that it can be agile, but also is capable of carrying an additional 40g of payload which can be used for sensors and processing to develop autonomy, which is significantly larger than other comparably sized platforms. In contrast, the comparably sized DelFly Explorer has a wingspan of 28cm and weighs 20g including autonomy electronics and the Purdue Hummingbird, which has a wingspan of 17cm weighs 12g.

The larger payload budget on Aerobat opens up the possibility of pushing the envelope for onboard perception and state estimation in flapping wing systems. Perception and state estimation in flapping wing robots is severely limited by the amount of computation possible onboard, and there are a limited number of works that have successfully demonstrated any level of onboard autonomy. \cite{garcia_bermudez_optical_2009}, \cite{mcguire_efficient_2017} and \cite{duhamel_altitude_2012} use optical flow for low-level control. Of these, only \cite{mcguire_efficient_2017} and \cite{garcia_bermudez_optical_2009} perform the computations onboard. \cite{de_wagter_autonomous_2014} uses a stereo rig to perform obstacle avoidance using onboard computation of disparity maps. The authors demonstrate autonomous avoidance of pillars during flight, but this method would struggle in more unstructured environments where depth information about obstacles needs to be more precise. \cite{tu_acting_2019} exploits its soft deformable wings by using them as sensors to detect wall collisions to navigate through a confined space. All of these works carry out onboard computation on small micro-controllers that can only handle basic autonomy. However, Aerobat's larger payload can support better processing, giving the opportunity to attempt more state of the art approaches. 

The state of the art in aerial robot perception has been established largely using multi-rotor platforms or offboard computation. Specifically, visual inertial approaches have gained much popularity due to cameras being cheap, lightweight and easily available, and complementing the noisy but high rate inertial data provided by an IMU. These have been implemented with variations in the type of visual data used (feature based \cite{campos_orb-slam3_2021, qin_vins-mono_2018} or direct \cite{engel_lsd-slam_2014, bloesch_iterated_2017}), the number of data points considered (full history, sliding window, latest only), methods for matching and estimation, back-end optimization \cite{kaess_isam2_2012}, loop closures, etc. and have been tested in various multi-rotor applications \cite{rhodes_autonomous_2022, weinstein_visual_2018}.

Considering the popularity and versatility of visual inertial odometry, this is chosen as the approach of choice for Aerobat. However, these algorithms typically require heavy processors to run in real time. Chapter \ref{chap:perception} Section \ref{subsec:processor} provides more details about the selection of processor on Aerobat and a comparison with processors typically used in these applications, however, here it is sufficient to say that care must be taken in selecting the visual inertial algorithm to be implemented on Aerobat. 

\begin{figure}
    \centering
    \includegraphics[width=0.5\textwidth]{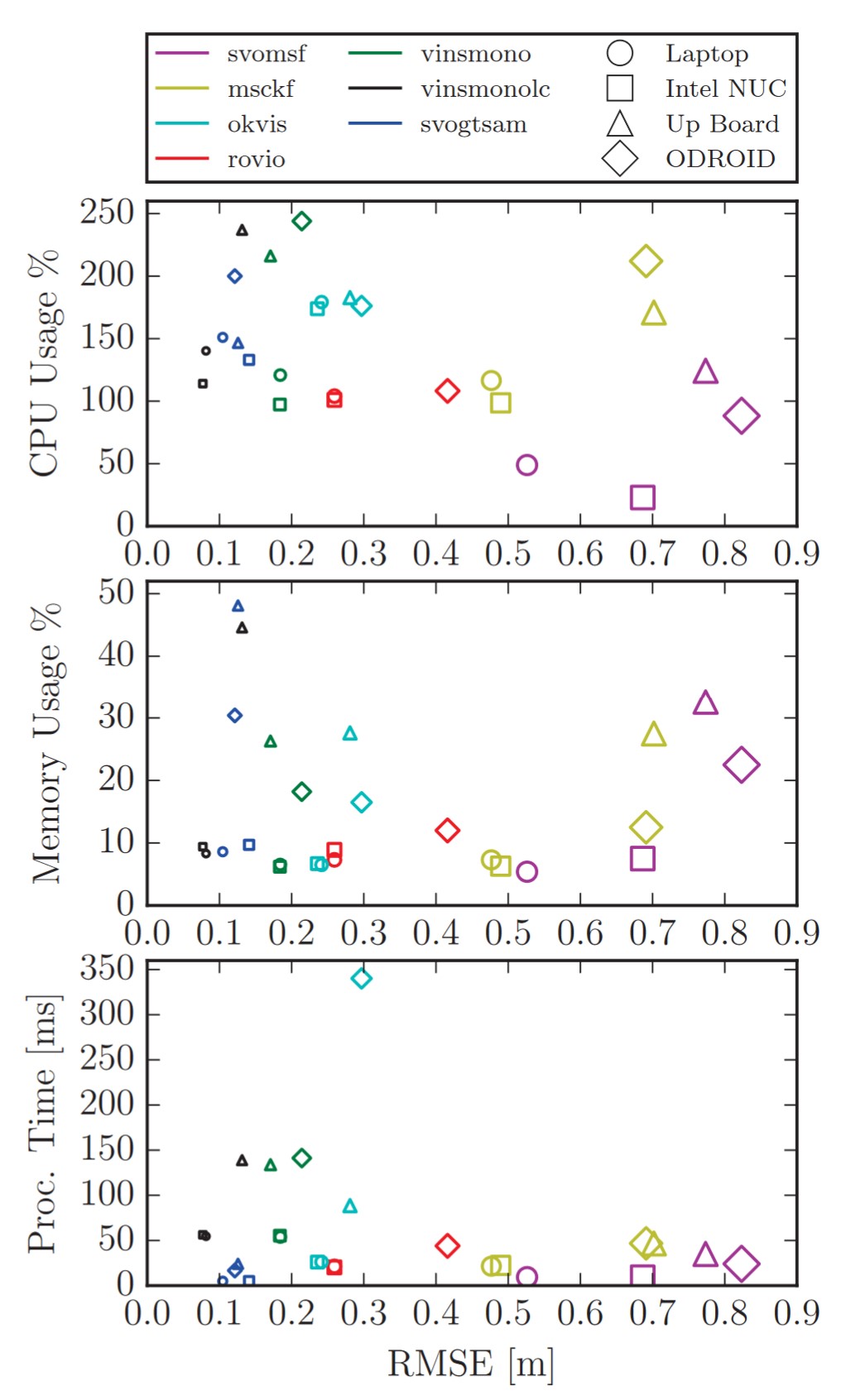}
    \caption{Figure from \cite{delmerico_benchmark_2018} depicting the performance of various VIO algorithms on different processors}
    \label{fig:benchmark}
\end{figure}

\cite{delmerico_benchmark_2018} compares the performance of visual inertial odometry algorithms on various processors, noting the amount of CPU and RAM usage, processing time and accuracy. Figure \ref{fig:benchmark}, from \cite{delmerico_benchmark_2018} shows the graph of performance of different Visual Inertial Odometry (VIO) algorithms. Of the processors compared in this work, Odroid XU4 and Intel Up Board relevant for comparison with Aerobat. With 2GB and 4GB of available RAM respectfully, they are on the lower end of typically used processors in these applications. More details about the two and a comparison is presented in Section \ref{subsec:processor}, but the results of this paper indicate that although there is a drop off in accuracy, lighter algorithms such as Multi-State Constrained Kalman Filter (MSCKF) and Semi-direct Visual-inertial Odometry + Multi-Sensor Fusion (SVOMSF) can run on limited hardware. Heavier algorithms such as SVO+GTSAM, OKVIS and ROVIO have larger processing times and consume more memory, but are also more accurate, and may potentially be implemented if the processing capacity of Aerobat is improved.

This provides hope that with further optimization and tailoring of these and newer algorithms to Aerobat's specific application, it is possible to run these state of the art algorithms close to real time on limited hardware.
\chapter{Towards untethered flight}
\label{chap:untethered}

Northeastern University's Aerobat is a project in development since 2016. \cite{sihite_mechanism_2020, sihite_enforcing_2020, hoff_optimizing_2018, hoff_reducing_2017, ramezani_biomimetic_2017} describe the development of the mechanical structure and actuation mechanism. \cite{sihite_unsteady_2022, ghanem_efficient_2021} describe the development of simulation models and trajectory planning. \cite{sihite_unsteady_2022} achieved tethered hovering flight indoors using these models on our indoor tethered test platform Aerobat Gamma.

The next stage of development was focused towards developing a second version of Aerobat, called Aerobat Beta for testing untethered flight outdoors.

\begin{figure}
    \centering
    \includegraphics[width=\textwidth, keepaspectratio]{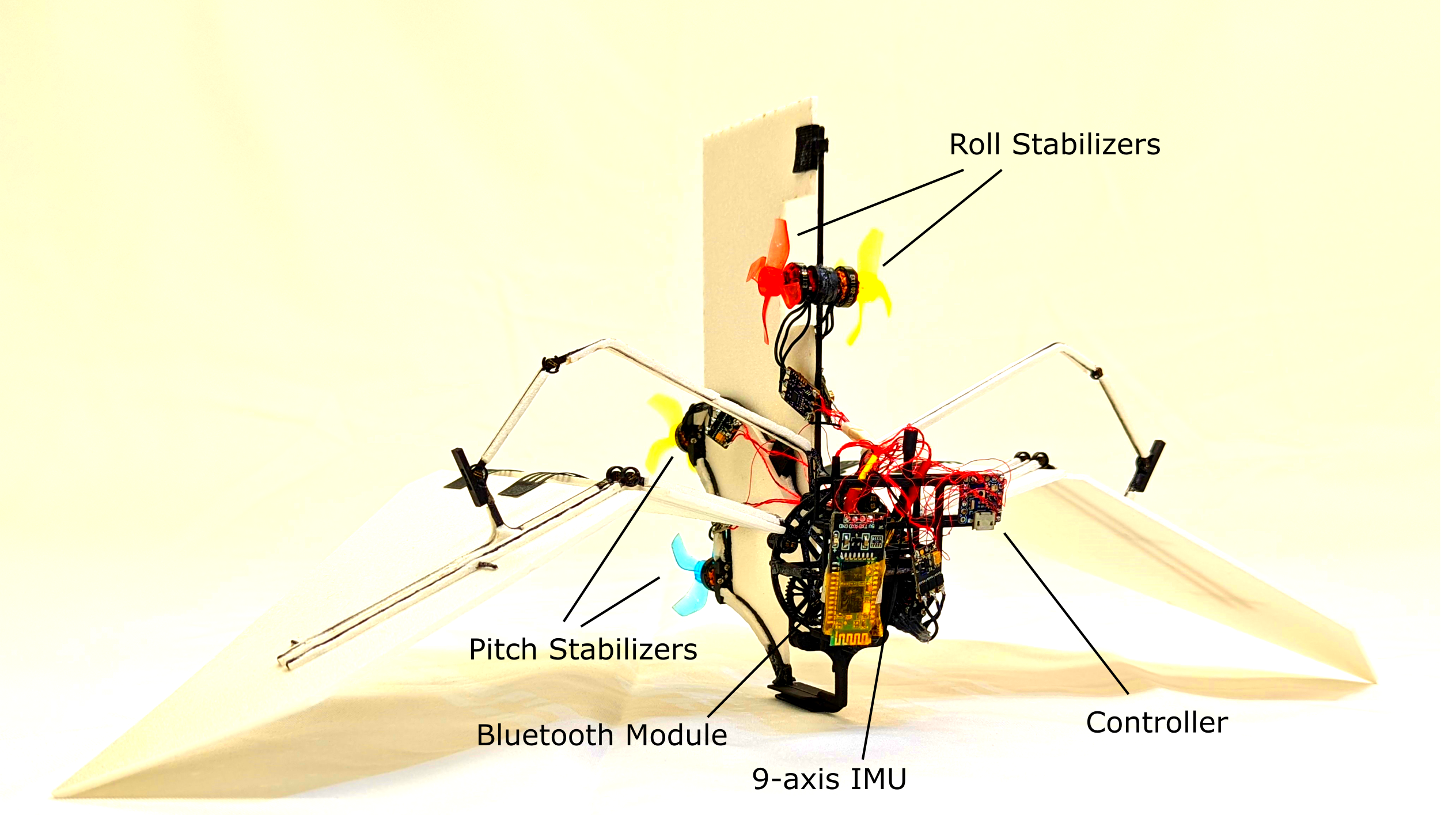}
    \caption{Aerobat Beta with onboard stabilizers}
    \label{fig:beta_old}
\end{figure}

Aerobat Beta was designed and built as part of an earlier Master's thesis presented in \cite{hu_bang-bang_2022}. As a test platform, Aerobat Beta has lightweight laser-cut foam wings that are easily replaceable. The original goal of Aerobat Beta was to test lift-generation capabilities in isolation, and to that end, stabilizers were added to stabilize the roll and pitch axes in flight, allowing the wings to generate lift based on an open loop PWM signal sent to the motor. Stabilization was carried out using a simple PD controller that read acceleration and gyroscope values from an onboard IMU to calculate roll and pitch. PWM signal data and IMU data were relayed to a ground-station computer over Bluetooth for debugging. All this was controlled onboard by an Arduino Pico micro-controller weighing 1g, with 24kB of memory.

\begin{figure}
    \centering
    \includegraphics[width=\textwidth]{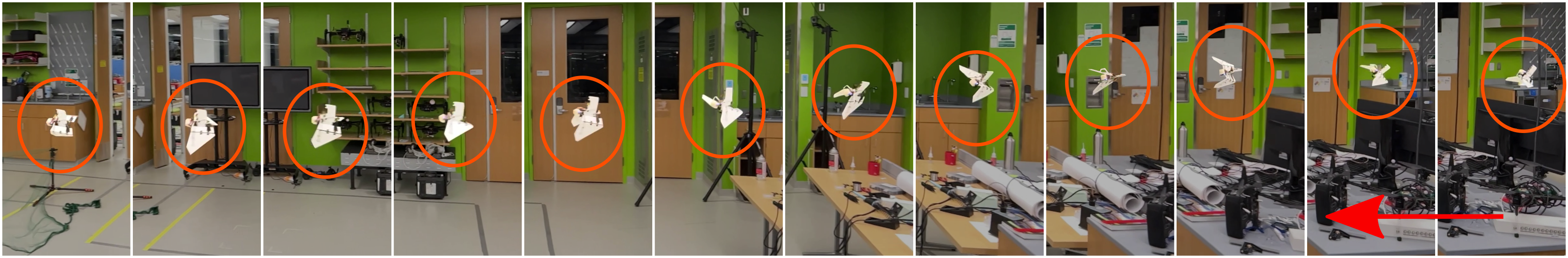}
    \caption{Snapshots of successful indoor 7m untethered flight showing stabilization in flight}
    \label{fig:indoor-untethered}
\end{figure}

At the start of the work presented in this thesis, Aerobat Beta was flying with intermittent success over short 3-5m distances. Testing was carried out indoors and the main focus was on increasing consistency of flight. The primary source of flight inconsistencies was found stem from the gear mechanism that keeps the two wings in sync. Additional inconsistencies came from poorly calibrated ESCs for the stabilizers and imbalanced weight. After strengthening 3D printed parts, cleaning up the gear mechanism and calibrating the ESCs, more consistent flight was observed, until finally 5-7m untethered flight was consistently achievable indoors. Figure \ref{fig:indoor-untethered} shows one such flight. The snapshots show untethered flight before Aerobat hits the safety net, showing orientation correction in the process. 

The modifications that allowed this to happen served only as temporary fixes and necessitated constant maintenance of the hardware to keep Aerobat in fly-worthy condition. As an early test platform, however, this was acceptable at the time and testing was continued. With consistent flight demonstrated indoors, Aerobat was taken outdoors for longer distance flights than could be executed in the indoor space available. 

\begin{figure}
    \centering
    \includegraphics[width=\textwidth]{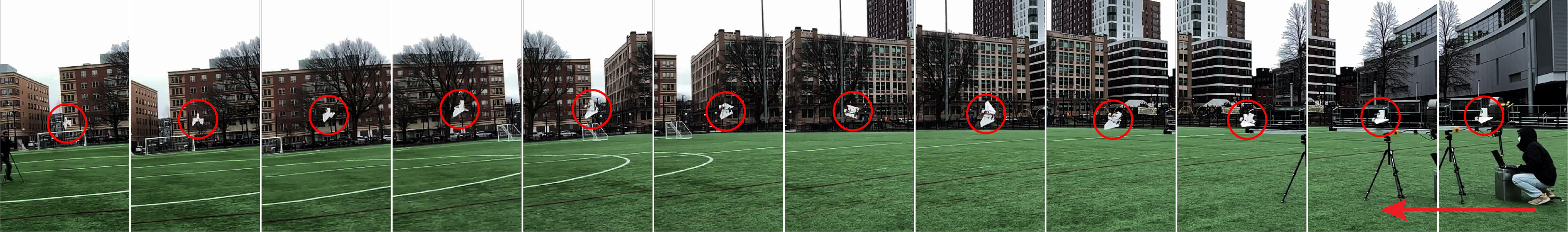}
    \caption{Snapshots of successful 10m outdoor untethered flight showing stabilization in flight}
    \label{fig:outdoor-untethered}
\end{figure}

Outdoor tests pose an additional challenge in the form of wind. Without closed loop control and only orientation based stabilization, testing can be difficult. However, with intermittent consistency, 10m outdoor flight was demonstrated. Figure \ref{fig:outdoor-untethered} shows one such flight, again showing Aerobat correcting undesired roll to continue flying.

This result sufficiently demonstrated lift generation capabilities of Aerobat Beta, and focus was shifted towards a long term fix for the gear mechanism and development of closed loop control. Chapter \ref{chap:control} describes the progress made towards development of closed-loop control. The rest of this chapter, however, will be dedicated to describing the development carried out to enable the work in Chapter \ref{chap:control} and beyond.

\section{Towards Safe Testing of Untethered Flight}
\label{sec:guard}

One of the issues faced while testing Aerobat outdoors was crashes. With foam wings and no protection, each crash would lead to large reset times, allowing for only a few tests to be conducted in a given time period. As more aspects of control are developed, the ability to perform multiple repeatable tests quickly will become very important. To this end, a guard design was proposed that would protect Aerobat in the event of crashes, reduce reset times and allow a large number of tests to be carried out.

\begin{figure}
    \centering
    \includegraphics[width=\textwidth]{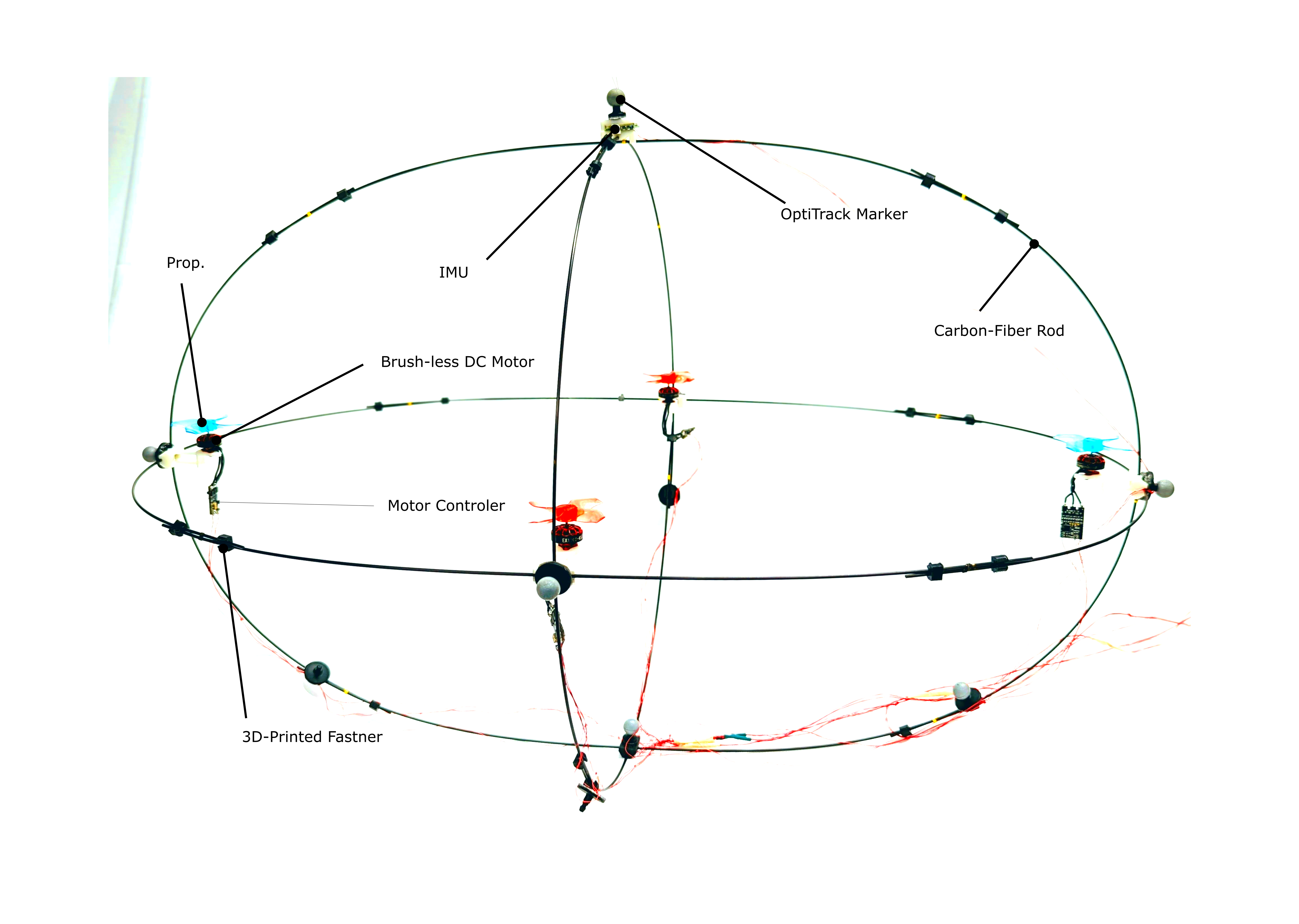}
    \caption{Shows guard design with stabilizers in place. Motors are currently placed on the inside for safe testing, but will eventually be moved to the outside}
    \label{fig:guard}
\end{figure}

\begin{figure}
    \centering
    \includegraphics[width=\textwidth, keepaspectratio]{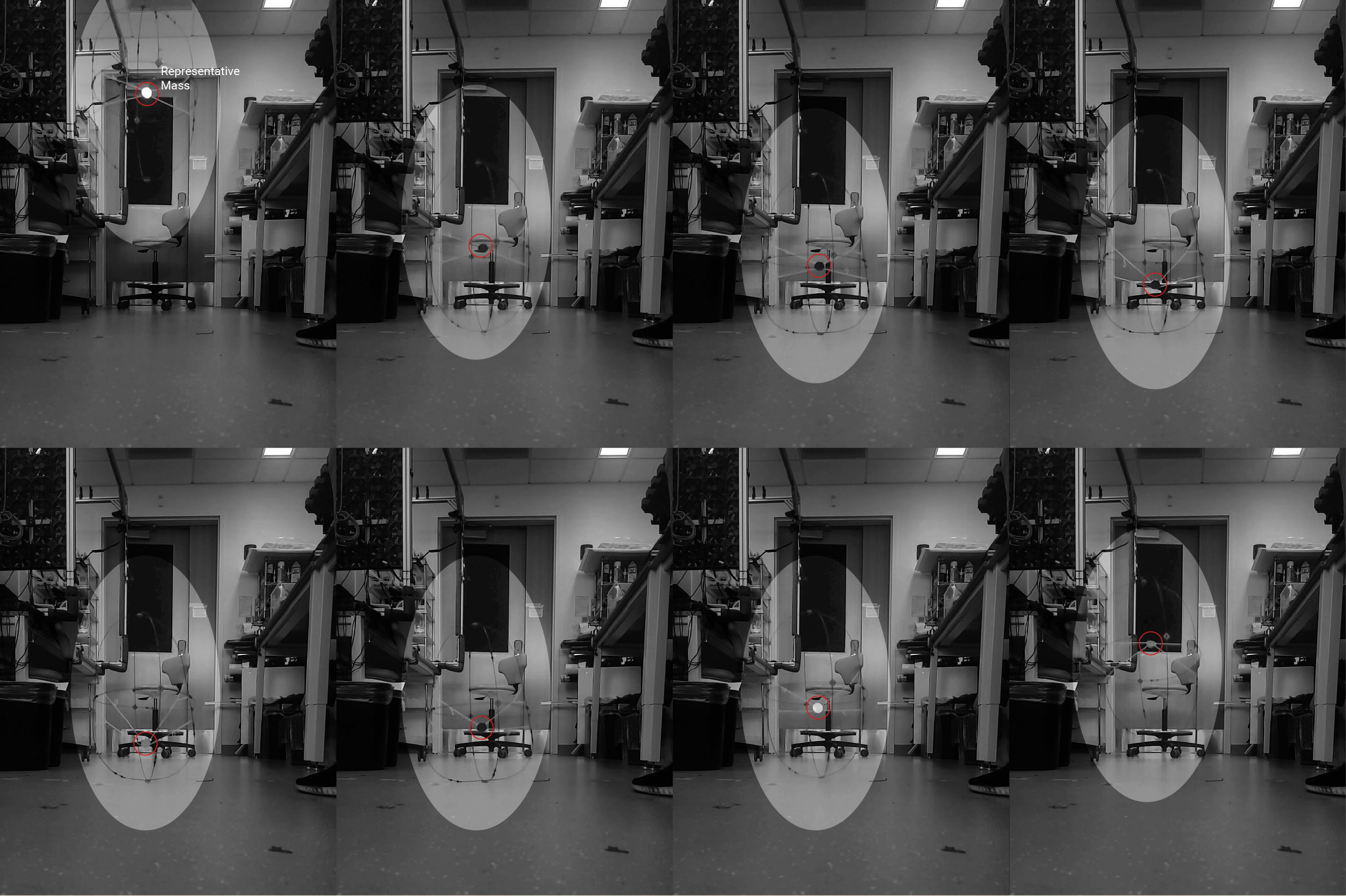}
    \caption{Shows drop test with guard}
    \label{fig:drop-test}
\end{figure}

Figure \ref{fig:guard} shows the proposed guard design with Aerobat mounted at the center. It has been named Kongming Lamp after the traditional Chinese lantern for it's distinctive shape and the safety it represents for Aerobat. Consisting of three concentric ellipses covering each of the three axes, this is designed to be a lightweight compliant addition that protects the robot in the event of a crash. Made of 11 lightweight carbon fiber rods, the structure provides strength and elasticity that would absorb impact in a crash. The rods are bound together by small snap-fit 3D printed parts that are optimized to reduce the weight to the minimum required. To test the strength of the guard, it was drop tested to see how a load at the center equivalent to the robot would survive. Figure \ref{fig:drop-test} shows the compliance of the structure absorbing the impact and protecting the representative weight.

An additional modification made in the interest of testing more advanced control is shifting the stabilizers from Aerobat to the guard and providing the guard with its own IMU. Having the guard independently stabilized isolates the robot from the guard dynamics and allows it to be used as much or as little as needed. Eventually, these stabilizers and the guard itself will be phased out and Aerobat will be robust enough to fly on its own. Figure \ref{fig:guard} shows the full guard design with stabilizers and IMU.

\begin{figure}
    \centering
    \includegraphics[width=0.75\textwidth]{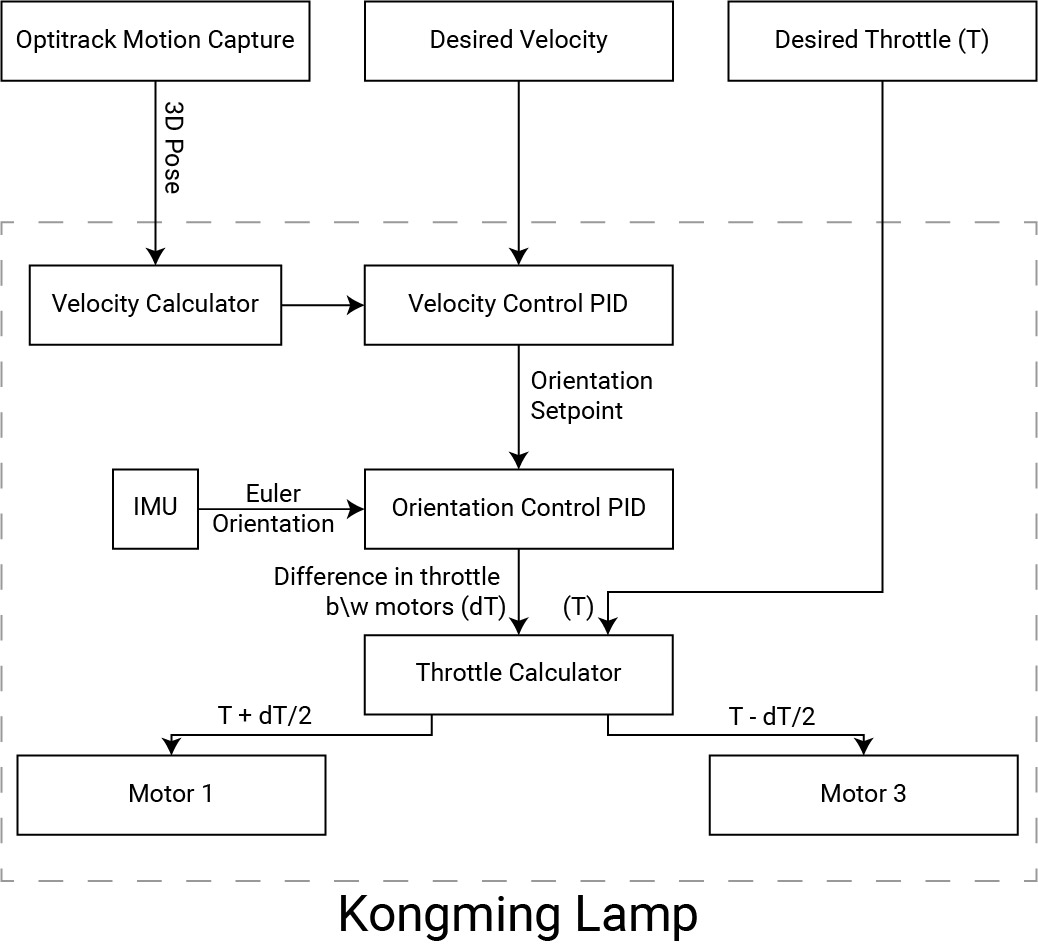}
    \caption{Logic diagram for guard controller. This is representative control logic for one axis. In the implementation, individual PID controllers are implemented for the x/pitch and y/roll axes.}
    \label{fig:guard_control}
\end{figure}

The guard is stabilized with the help of four BLDC motors arranged in a quad-copter-like configuration. The control algorithm for the guard runs on Aerobat's processor and uses feedback from its own IMU for independent control. Within RISE Arena, it is fitted with markers and tracked using Optitrack Motion Capture to provide pose information to the controller. Figure \ref{fig:guard_control} shows the controller logic used to stabilize the guard. For simplicity, only the roll and pitch orientations of the guard are stabilized, and velocity in only the x and y directions is considered. Altitude control will be part of future development.

Stabilizing the guard is challenging due to the compliant nature of the structure. The motors and IMU are mounted on snap-fit 3D printed parts that may slide along the carbon fiber rod. The rods themselves also stretch over time and the relative positioning between the motors is not rigid. This leads to challenges in tuning the controls for the guard as it needs to be robust enough to compensate for all these inconsistencies.

\section{Robotics-Inspired Study and Experimentation (RISE) Arena}
\label{sec:rise}

In order to develop controls for Aerobat, a fully controlled and repeatable environment is required where each aspect of Aerobat's dynamics may be isolated and individually studied. It needs a safe environment to test and tune controls in a rigorous and repeatable manner before it is ready to be taken outdoors for fully untethered flight.

\begin{figure}
    \centering
    \includegraphics[width=\textwidth]{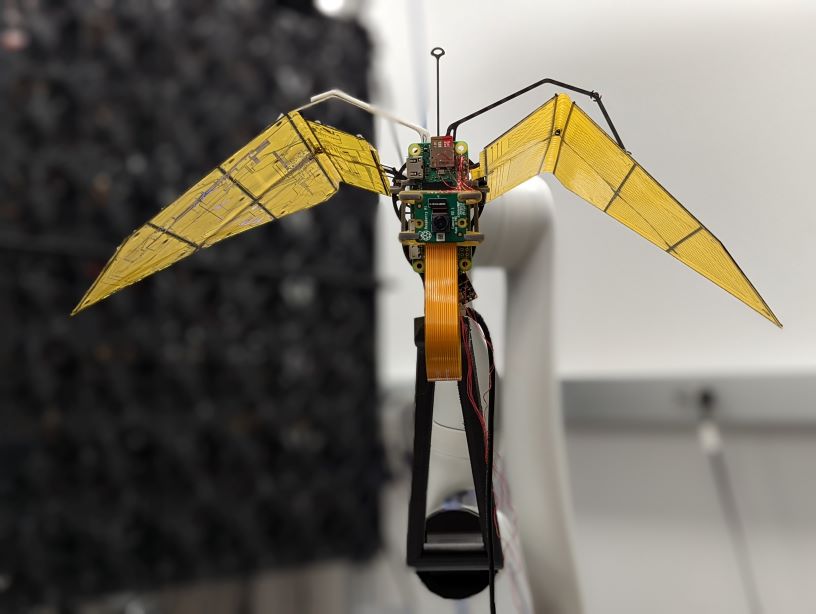}
    \caption{Aerobat Gamma}
    \label{fig:gamma}
\end{figure}

\begin{figure}
    \centering
    \includegraphics[width=\textwidth, keepaspectratio]{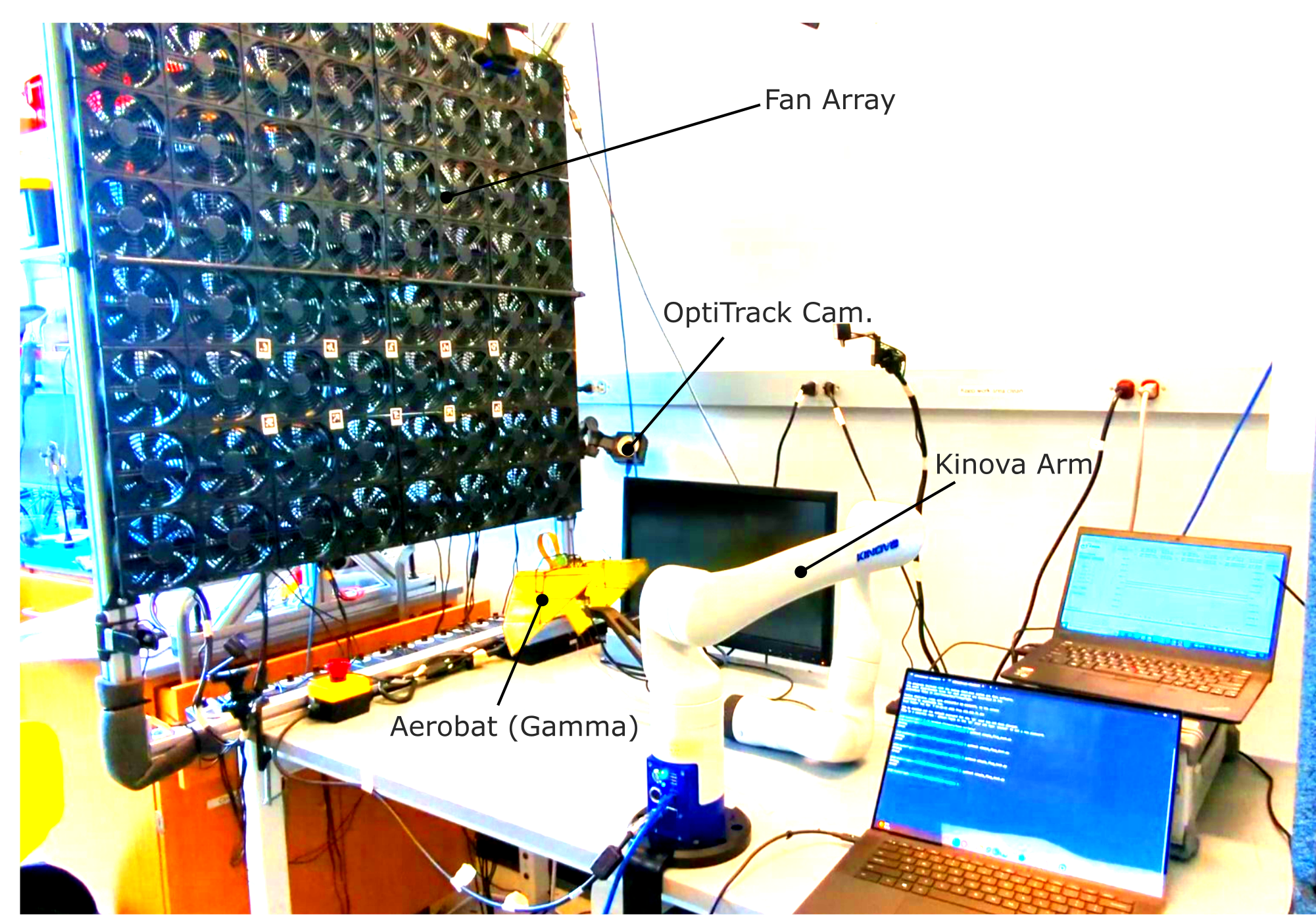}
    \caption{RISE Arena Components}
    \label{fig:rise}
\end{figure}

\begin{figure}
    \centering
    \includegraphics[width=0.8\textwidth]{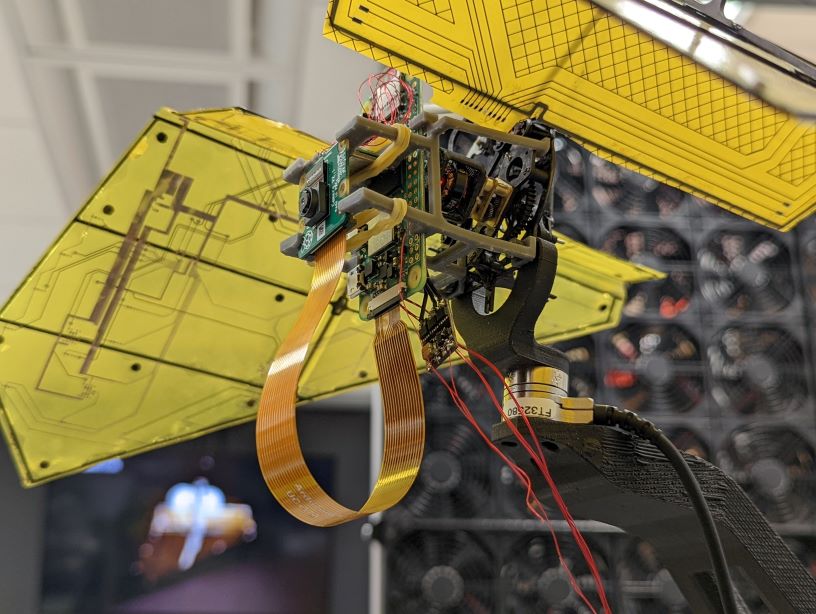}
    \caption{Integrated electronics components onto Aerobat Gamma}
    \label{fig:gamma_closeup}
\end{figure}

The Robotics-Inspired Study and Experimentation (RISE) Arena was created to provide this controlled test environment. Figure \ref{fig:rise} shows the setup of RISE Arena. At the center of it is the indoor tethered test platform Aerobat Gamma (Fig. \ref{fig:gamma}). Aerobat Gamma is a tethered version of Aerobat with flexible electronics in its wings. It is mounted on a highly sensitive ATI 6-axis load cell (shown in Fig. \ref{fig:gamma_closeup}). The robot and the load cell together are mounted at the end of a programmable 6 DOF manipulator. One side of RISE Arena is entirely covered by a large array of fans that can generate wind speeds of up to 2 m/s and the whole area is covered by 6 Optitrack Motion Capture Cameras.

\begin{figure}
    \centering
    \includegraphics[width=0.5\textwidth]{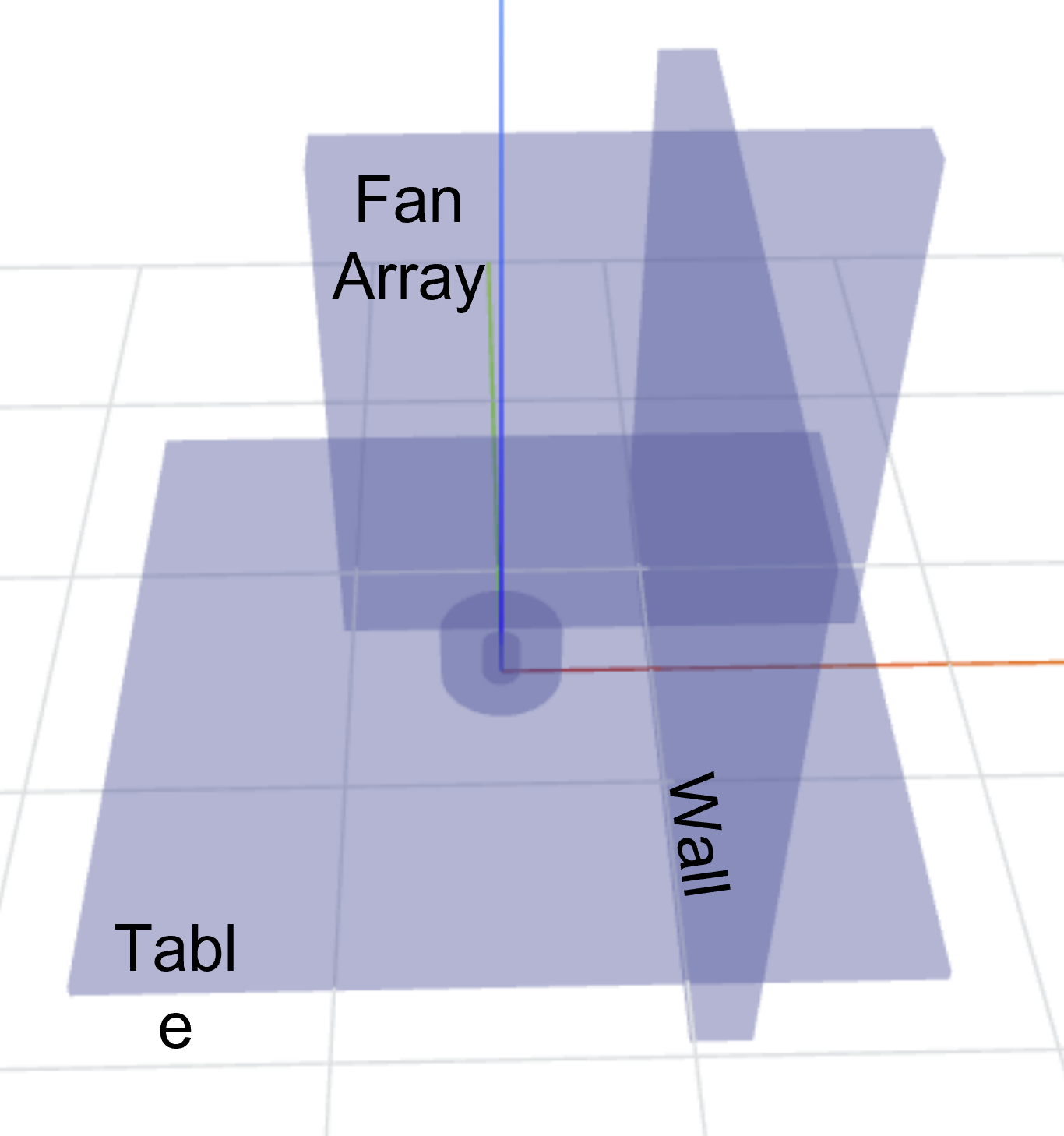}
    \caption{Protection Zones for RISE Arena}
    \label{fig:protection_zones}
\end{figure}

The robotic arm offers the ability to create trajectories with precise ground truth information available and do highly repeatable experiments. The arm is interfaced through Ethernet using a Python API. A wrapper was developed for the API that added new functionality, making it easier to interface with the arm, generate trajectories and execute predefined movements. Using the wrapper, keyboard teleoperation of the arm was developed, allowing a user to move the arm to any location and save the coordinates as waypoints in a trajectory. The waypoints may be saved and fed to different programs that execute different trajectories, controlling the duration and smoothness of the trajectory, and the number of loops of the trajectory to execute. It also enables setting protection zones (Fig. \ref{fig:protection_zones}) to protect the arm and the robot from collisions within RISE arena, allowing safe testing of controls.

From the motion capture cameras and the load cell, RISE Arena provides ground truth for flapping frequency, robot pose, lift generated, and aerodynamic forces on the robot, allowing controlled motion and pose within known stable wind conditions, making this a powerful tool for testing.

RISE Arena has been used throughout this work, from validating the the aerodynamic model to testing the guard controller to calibrating sensors and testing perception.

\section{Concluding remarks}

In this chapter, the preliminary results for untethered flight was presented with successful indoor and outdoor flight tests demonstrating a proof-of-concept for untethered flight. These flights were open loop. Future development will be focused towards developing closed loop control, with initial steps for this described in Chapter \ref{chap:control}. To better enable testing controls in closed loop flight, this chapter also describes the development of Kongming Lamp, a lightweight protective guard around Aerobat to save it from crashes and stabilize it while controls are being tuned. Finally, this chapter describes the development of indoor test setup RISE Arena, providing elaborate ground truth and a controlled repeatable environment for system identification and testing of controls. RISE Arena is far from a finished product, with many developments planned, including "free flight" of the robot while still attached to the manipulator using admittance control, incorporating more precise aerodynamic sensing and wind pattern detection and adding offboard processing to test more experimental and advanced algorithms. 
\chapter{Aerobat Modeling}
\label{chap:control}

This chapter describes the progress made towards developing a control model of Aerobat capable of executing trajectories. In order to do this, a model must be developed mapping between robot motion and control inputs to the actuators. \cite{sihite_unsteady_2022} makes progress towards this with a description of the aerodynamic model.

\begin{figure*}[t]
    \centering
    \vspace{0.1in}
    \includegraphics[width = 0.8 \linewidth]{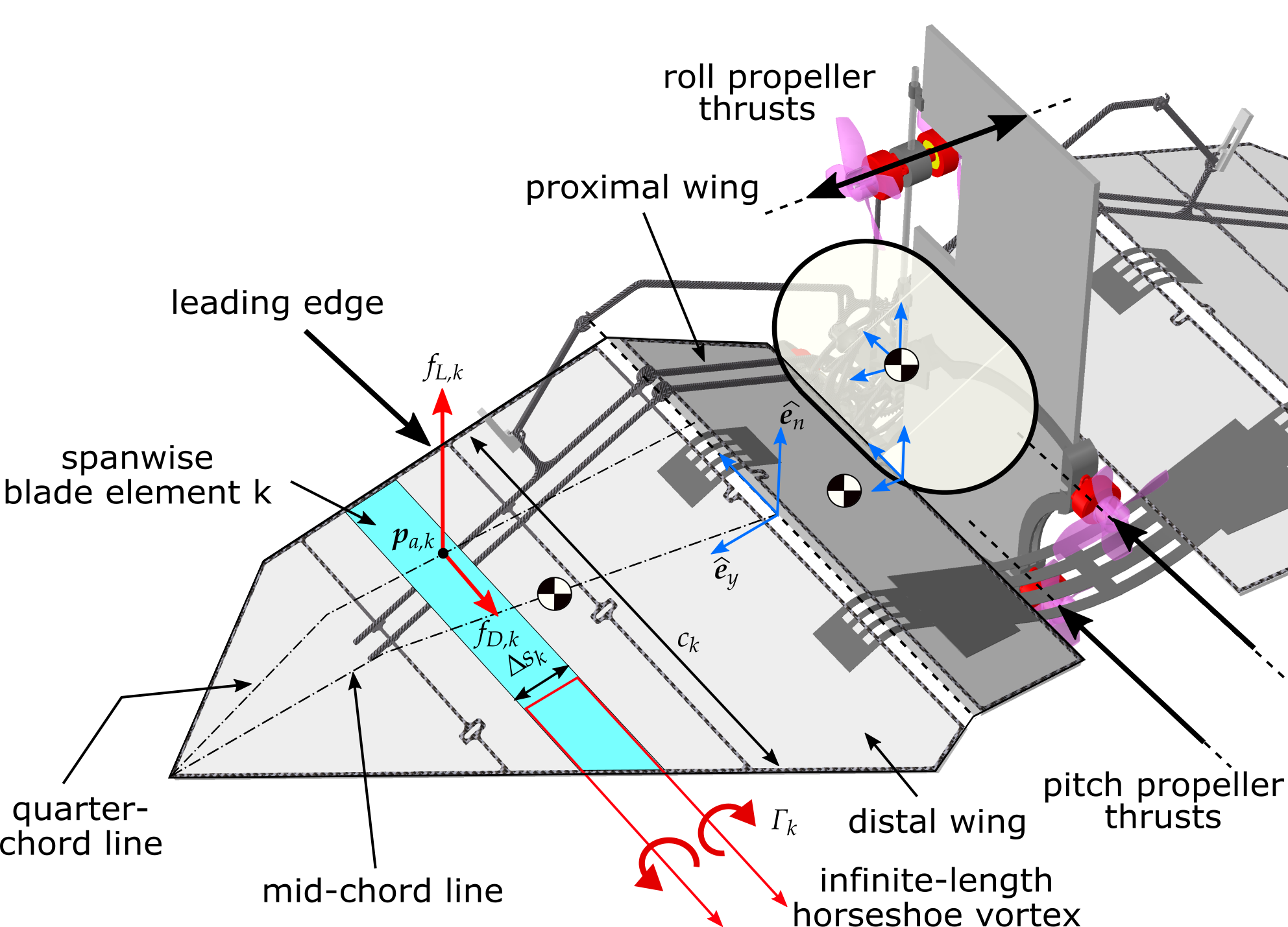}
    \caption{Free body diagram of Aerobat Left Wing}
    \vspace{-0.1in}
    \label{fig:fbd}
\end{figure*}

The dynamic modeling is derived using an unsteady aerodynamic model from the Wagner model and lifting-line theory \cite{boutet_unsteady_2018}. Aerobat has 20 degrees of freedom, but due to the nature of the kinetic sculpture of Aerobat's mechanism, this can be reduced to just 7 degrees of freedom (6 for the body and 1 for the motor that controls the flapping) with the rest expressed as kinematic constraints.

The dynamical equation of motion used in the simulation can be derived using Euler-Lagrangian dynamical formulations. Figure \ref{fig:fbd} shows the free-body diagram of the robot, which can be presented using 5 bodies: main body, proximal and distal wings of both sides. The synchronized wing trajectory allows us to just use one side of the wing in the states.

Let $\bm q = [\bm p^\top, \bm \theta^\top, q_s, q_e]^\top$ be the generalized coordinates, where $\bm p$ is the body center of mass inertial position, $\bm \theta$ is the Euler angles of the body, $q_s$ and $q_e$ are the left wing's shoulder and elbow angles, respectively. The dynamical equation of motion of the simplified system can be defined as follows:
\begin{equation}
\begin{aligned}
    \bm M(\bm q) \, \ddot{\bm q} &= \bm h(\bm q, \dot{\bm q}) + \bm u_a + \bm u_t + \bm J_c^\top \bm \lambda  \\
    \bm J_c \, \ddot{\bm q} &= [\ddot q_s, \ddot q_e]^\top = \bm y_{ks}, 
\end{aligned}
\label{eq:dynamic_eom}
\end{equation}

\noindent where $M$ is the inertial matrix, $\bm h$ is the gravitational and Coriolis forces, $\bm u_a$ and $\bm u_t$ are the generalized aerodynamic and thruster forces, respectively. $\bm \lambda$ is the Lagrangian multiplier which enforces the constraint forces acting on $q_s$ and $q_e$ to track the KS flapping acceleration $\bm y_{ks}$. $\bm \lambda$ can be solved algebraically from \ref{eq:dynamic_eom} given the states $\bm x = [\bm q^\top, \dot{\bm q}^\top]^\top$ and both generalized forces $\bm u_a$ and $\bm u_t$. These generalized forces can be derived using virtual displacement, as follows:

\begin{equation}
\begin{aligned}
    \bm u_a &= \sum_{i=1}^{N_b} B_{a,i}(\bm q)\, \bm f_{a,i} \quad &
    \bm u_t &= \sum_{i=1}^{N_t} B_{t,i}(\bm q)\, \bm f_{t,i}
\end{aligned}
\label{eq:generalized_forces}
\end{equation}
where $B$ matrices map the forces $\bm f \in \mathbb{R}^3$ to the generalized coordinates $\bm q$, $N_b$ is  the number of blade elements, and $N_b$ is the number of thrusters. Let the position $\bm p_k(\bm q)$ be the inertial position where the force $\bm f_k$ defined in the inertial frame is applied. The matrix $B_k$ for this force can be derived as follows: $B_k = \left( \partial \dot{\bm p}_{k} / \partial \dot{\bm q} \right)^\top$. The aerodynamic forces generated on each blade elements and thrust forces are combined to form $\bm u_a$ and $\bm u_t$, respectively.

The aerodynamics can be derived using discrete blade elements following the derivations in \cite{boutet_unsteady_2018}. This model uses the lifting line theory and Wagner's function to develop a model for calculating the lift coefficient. Let $S$ be the total wingspan and $y \in [-S/2, S/2]$ represents a position along the wingspan. The vortex shedding distribution can be defined as a function of truncated Fourier series of size $m$ across the wingspan, as follows:
\begin{equation}
\begin{gathered}
    \Gamma(t,y) = \frac{1}{2} a_0 \, c_0 \, U \, \sum^{m}_{n=1} a_n(t) \, \sin(n\,\theta(y))
\end{gathered}
\end{equation}
\noindent where $a_n$ is the Fourier coefficients, $a_0$ is the slope of the angle of attack, $c_0$ is the chord length at wing's axis of symmetry, and $U$ is the free stream airspeed. Let $\theta$ be the change of variable defined by $y = (S/2)\cos(\theta)$ for describing a position along the wingspan $y \in (-S/2, S/2)$. From $\Gamma(t,y)$, we can derive the additional downwash induced by the vortices, defined as follows:
\begin{equation}
\begin{aligned}
    w_{y}(t,y) &
    = - \frac{a_0 c_0 U}{4S} \sum^{m}_{n=1} n a_n(t)  \frac{\sin(n \theta)}{\sin(\theta)}.
\end{aligned}
\label{eq:induced_downwash}
\end{equation}

Following the unsteady Kutta-Joukowski theorem, the sectional lift coefficient can be expressed as follows:
\begin{equation}
\begin{aligned}
    C_L(t,y) &= a_0 \sum^{m}_{n=1} \left( \frac{c_0}{c(y)} a_n(t) + \frac{c_0}{U} \dot{a}_n(t) \right) \sin(n\theta),
\end{aligned}
\label{eq:lift_coeff_fourier}
\end{equation}
where $c(y)$ is the chord length at the wingspan position $y$. The computation of the sectional lift coefficient response of an airfoil undergoing a step change in downwash $\Delta w(y) << U$ can be expressed using Wagner function $\Phi(t)$:
\begin{equation}
\begin{aligned}
    c_L(t,y) &= \frac{a_0}{U} \Delta w(t,y) \Phi(\tilde t) \\
    \Phi(\tilde t)    &= 1 - \psi_1 e^{-\epsilon_1 \tilde t} - \psi_2 e^{-\epsilon_2 \tilde t}
\end{aligned}
\label{eq:lift_coeff_wagner}
\end{equation}

where $\tilde t(t) = \int_0^t (v_e^i/b) dt$ is the normalized time which is defined as the distance traveled divided by half chord length ($b = c/2$). Here, $v_e^i$ is defined as the velocity of the quarter chord distance from the leading edge in the direction perpendicular to the wing sweep. For the condition where the freestream airflow dominates $v_e$, then we can approximate the normalized time as $\tilde t = Ut/b$. The Wagner model in \eqref{eq:lift_coeff_wagner} uses Jones' approximation \cite{boutet_unsteady_2018}, with the following coefficients: $\psi_1 = 0.165$, $\psi_2 = 0.335$, $\epsilon_1 = 0.0455$, and $\epsilon_2 = 0.3$.

Duhamel's principles can be used to superimpose the transient response due to a step change in downwash as defined in \eqref{eq:lift_coeff_wagner}. Additionally, integration by parts can be used to simplify the equation further, resulting in the following equation:
\begin{equation}
\begin{aligned}
    C_L(t,y) &= \frac{a_0}{U} \left( w(t,y) \Phi(0) - \int_{0}^{t} \frac{\partial \Phi(t - \tau)}{\partial \tau} w(\tau, y) d\tau \right).
\end{aligned}
\label{eq:aero_CL_base}
\end{equation}
\begin{equation}
\begin{aligned}
    \frac{\partial \Phi(t - \tau)}{\partial \tau} &=
    -\frac{\psi_1 \epsilon_1 U}{b} e^{-\frac{\epsilon_1 U}{b}(t-\tau)}
    -\frac{\psi_2 \epsilon_2 U}{b} e^{-\frac{\epsilon_2 U}{b}(t-\tau)}
\end{aligned}
\label{eq:partial_phi}
\end{equation}
Here, $w(t,y)$ is the total downwash defined as:
\begin{equation}
    w(t,y) = v_n(t,y) + w_y(t,y),
\label{eq:total_downwash}
\end{equation}
where $v_n$ is the airfoil velocity normal to the wing surface which depends on the freestream velocity and the inertial dynamics. Finally, we can represent the integrals as the following states:
\begin{equation}
\begin{aligned}
    z_{1} (t,y) &= \int_{0}^{t} \frac{\psi_1 \epsilon_1 U}{b} e^{-\frac{\epsilon_1 U}{b}(t-\tau)} w(\tau,y) d\tau
    \\
    z_{2} (t,y) &= \int_{0}^{t} \frac{\psi_2 \epsilon_2 U}{b} e^{-\frac{\epsilon_2 
    U}{b}(t-\tau)} w(\tau,y) d\tau.
\end{aligned}
\label{eq:aero_states_z}
\end{equation}
Both of these states can be expressed as an ODE by deriving the time derivatives of \eqref{eq:aero_states_z}. They can be derived using Leibniz integral rule, yielding the following equations:
\begin{equation}
\begin{aligned}
    \dot z_{1} (t,y) &= \frac{\psi_1 \epsilon_1 U}{b} \left( w(t,y) - \frac{\epsilon_1 U}{b} z_1(t,y) \right) \\
    \dot z_{2} (t,y) &= \frac{\psi_2 \epsilon_2 U}{b} \left( w(t,y) - \frac{\epsilon_2 U}{b} z_2(t,y) \right).
\end{aligned}
\label{eq:aero_states_dz}
\end{equation}
The sectional lift coefficient can then be defined as:
\begin{equation}
\begin{aligned}
    c_L(t,y)  = \frac{a_0}{U} \left( w(t,y) \phi(0) + z_1(t,y) + z_2(t,y) \right),
\end{aligned}
\label{eq:aero_CL_final}
\end{equation}
and we can march the aerodynamic states $z_1$ and $z_2$ forward in time using \eqref{eq:aero_states_dz}. Finally, we can relate the both sectional lift coefficient equations in \eqref{eq:lift_coeff_fourier} and \eqref{eq:aero_CL_final} to solve for the Fourier coefficient rate of change, $\dot{a}_n$. 

The aerodynamic states are defined along the span of the wing and can be discretized into $m$ blade elements. Therefore, we can derive the $m$ equations relating \eqref{eq:lift_coeff_fourier} and \eqref{eq:aero_CL_final} on each blade element to solve for the $\dot{a}_n$. Then, including $z_1$ and $z_2$ on each blade elements, we will have $3m$ ODE equations to solve. We can represent $a_n$, $z_1$, and $z_2$ of all blade elements as the vector $\bm a_n \in \mathbb{R}^{m}$, $\bm {z}_1 \in \mathbb{R}^{m}$, and $\bm z_2 \in \mathbb{R}^{m}$, respectively. 

\begin{figure}
    \centering
    \includegraphics[width=\textwidth, keepaspectratio]{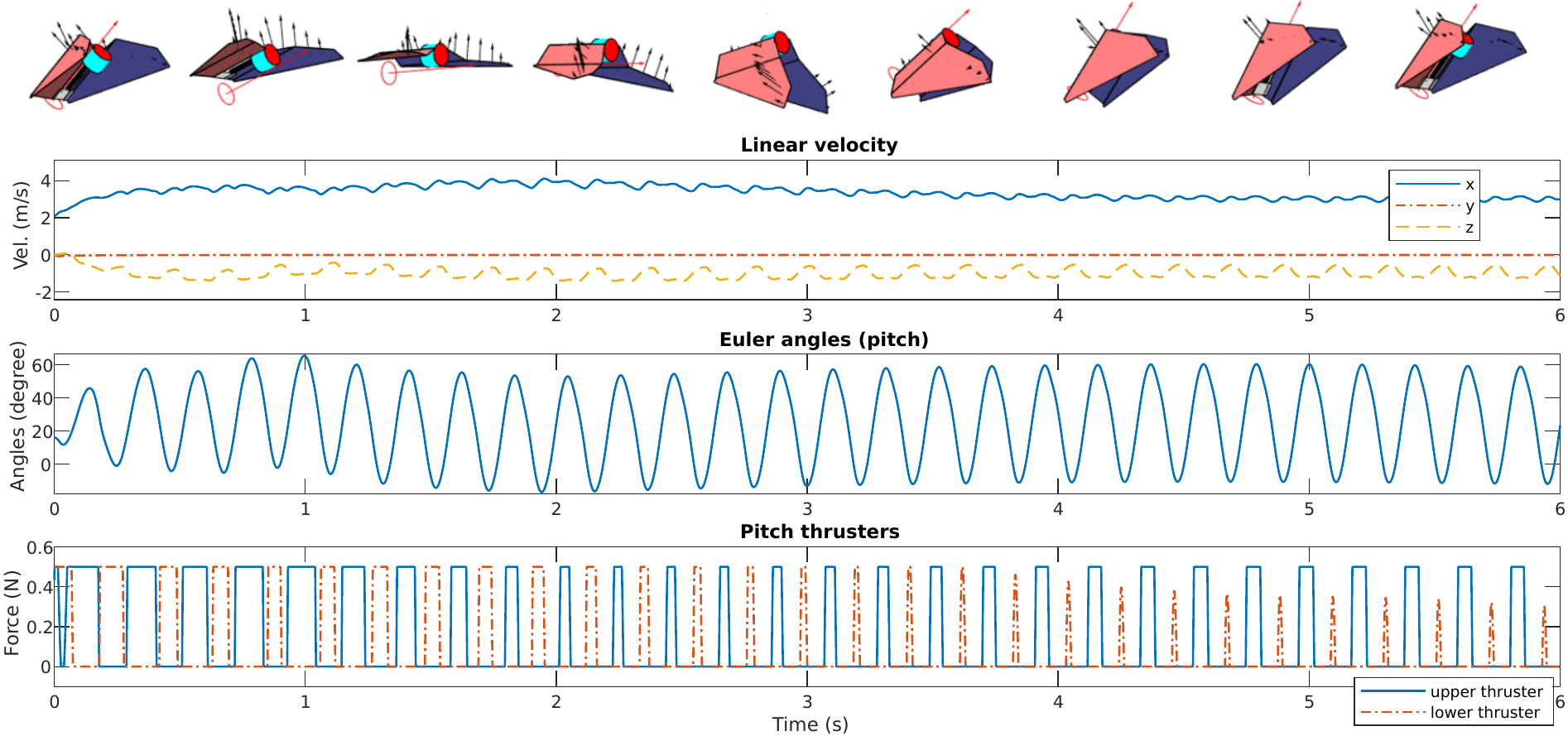}
    \caption{Illustrates Aerobat's stick-diagram and simulated state trajectories under bang-bang control of the longitudinal and frontal dynamics.}
    \label{fig:simulation_results}
\end{figure}

This model was simulated in \cite{sihite_bang-bang_2022} (Fig. \ref{fig:simulation_results}) and partially validated by the IMU data from untethered flight tests . However, to fully validate the model and close the loop, a more controlled testing setup is required.

\section{Validation of Aerodynamic Model}

\begin{figure}
    \centering
    \includegraphics[width=0.8\textwidth, keepaspectratio]{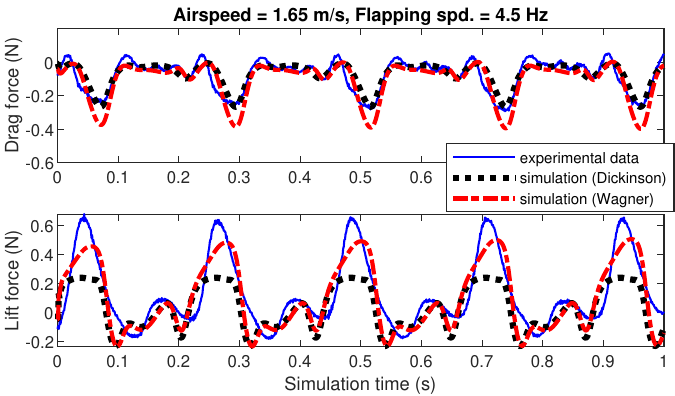}
    \caption{Load cell experimental data (solid line) vs. the simulated lift and drag generated by the quasi-steady Dickinson's model (dotted line) and Wagner aerodynamic model (dashed line). The robot was subjected to various airspeed and flapping frequencies, which was then simulated using the model derived under the same conditions.}
    \label{fig:validation}
\end{figure}

Using RISE Arena, the aerodynamic model presented in \cite{sihite_unsteady_2022} was validated. Aerobat was set to flap at a fixed known frequency of about 2 Hz and load cell measurements were taken for headwind speeds of 0.5, 1.0, and 1.5 m/s. The results closely match the simulation, validating this model (Fig. \ref{fig:validation}) .

\section{Concluding Remarks}

This chapter presented the aerodynamic model of Aerobat and the steps taken towards validating it using the newly setup RISE Arena (Section \ref{sec:rise}), taking Aerobat one step closer to closed-loop control. Future development will be focused towards system identification and addition of more degrees of actuation into the wings, allowing Aerobat to control roll and pitch dynamics. 

\chapter{Perception Challenges and Preliminary Works}
\label{chap:perception}

As described in the introduction (Chap. \ref{chap:introduction}), Aerobat needs both high level and low level control in order to execute autonomous flight. This chapter describes the progress made towards developing perception and state estimation onboard Aerobat, from selecting the electronics required to hardware and software integration and sensor calibration. 

\section{Onboard Electronics}
\label{sec:electronics}

When selecting the onboard electronics for Aerobat, a soft payload limit of 15g was imposed on the selection for autonomy electronics. This was done to allow for additional stabilizers used for testing in the initial stages of development while the controls are still under research.

\subsection{Processor}
\label{subsec:processor}

In order to achieve autonomy, the onboard processor must be powerful enough to interface with multiple sensors and execute control algorithms at a high enough rate. With a 15g payload limit, this did not offer a lot of options, forcing a compromise between processing power and weight.

Multirotors have a larger payload capacity and can carry large processors. \cite{tang_aggressive_2018, foehn_agilicious_2022, kaufmann_beauty_2019, chen_aerial_2022} use Odroid XU4 and Intel UP board onboard, both of which weigh around 40g and come with 4-core ~2GHz 64-bit processors with 2/4GB of RAM. Some such as \cite{petrlik_robust_2020, best_resilient_2022, foehn_alphapilot_2022, foehn_time-optimal_2021} use even larger processors such as Intel NUC (9 cores, 1.1GHz, 16GB RAM) or one of the NVIDIA Jetson series: Nano (4-core CPU, 128-core GPU, 4GB RAM), TX2 (2-core CPU, 256-core GPU, 4GB RAM), Xavier (8-core CPU, 512-core GPU, 32 GB RAM), which all weigh in the order of a few 100g. All these options are far beyond the payload capacity of Aerobat.

At the other extreme, small microcontroller boards such as Arduino Nano, Arduino Pico, and Raspberry Pi Pico are becoming more capable. Arduino Pico (weighing just 1g) was used in the initial stages of Aerobat flight tests to control the actuator and stabilizer motors (Fig. \ref{fig:beta_old}). However, all of these have under 1MB of memory and are not practical for high-level computation. 

Arduino Portenta is a small 2-core microcontroller board that can simultaneously run an Arduino loop on one core and computer vision algorithms on the other, with support for TensorFlow Lite. Arduino Nicla Vision is another lightweight microcontroller board option that has a camera, IMU, and distance sensor embedded in the board itself and supports TinyML, OpenMV, and MicroPython. Both Arduino Portenta and Nicla come with Bluetooth and WiFi embedded.

These are potentially attractive options for specialized applications. However, Portenta has just 8MB of memory (expandable up to 64MB) and Nicla is even lower at just 2MB, which is not sufficient for the level of autonomous computation targeted for Aerobat. 

\begin{figure}
    \centering
    \begin{subfigure}[b]{0.3\textwidth}
        \includegraphics[width=\textwidth, keepaspectratio]{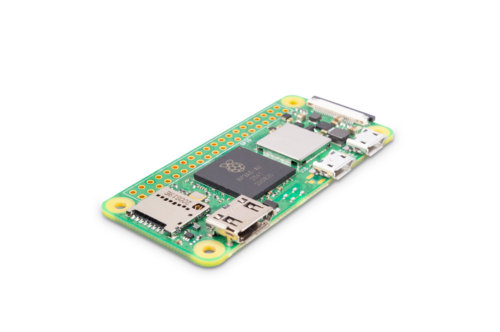}
        \caption{The RPi Zero 2 W}
        \label{fig:pizero}
    \end{subfigure}
    \hfill
    \begin{subfigure}[b]{0.3\textwidth}
        \includegraphics[width=0.5\textwidth, keepaspectratio]{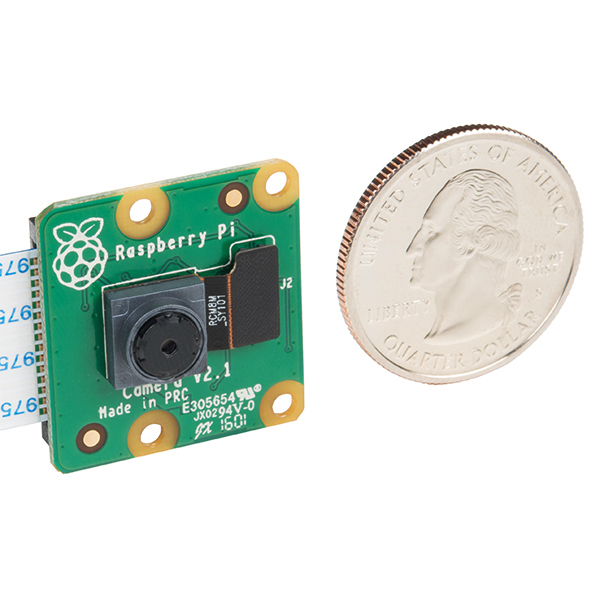}
        \caption{RPi Camera Module}
        \label{fig:rpicamera}
    \end{subfigure}
    \hfill
    \begin{subfigure}[b]{0.3\textwidth}
        \includegraphics[width=0.5\textwidth, keepaspectratio]{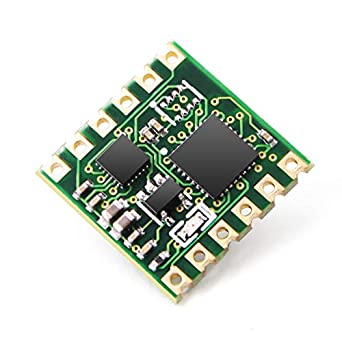}
        \caption{WT901 9 Axis IMU}
        \label{fig:imu}
    \end{subfigure}
    \caption{Onboard electronic components}
    \label{fig:onboard_electronics_photos}
\end{figure}

After some consideration, the Raspberry Pi Zero 2 W (Fig. \ref{fig:pizero}) was chosen as the ideal compromise. Weighing 11g, the Raspberry Pi Zero 2 W runs on a 4-core 1GHz 64-bit ARM processor with a Linux-based operating system. It has 512MB of RAM, Wi-Fi, and Bluetooth capability, and has an interface for a Raspberry Pi camera. This is still relatively powerful for its size, and at the time of writing to the best of my knowledge, is the most powerful processor weighing less than 15g available on the market.

\subsection{Sensors}

The choice of onboard sensors depends on the approach used for autonomy and the environment in which it is designed to operate. For example, an IR camera might be suitable for dark environments such as night flight, and simple laser rangefinders might suffice for maintaining a steady heading within a confined space such as a tunnel. Other options considered included Sonar rangefinders, optical flow sensors, and stereo cameras. 

At this early development stage, however, priority is given to versatility that would allow for testing under a range of environments and applications. With this in mind, a single monocular RGB camera and IMU were chosen as the onboard sensors, using a visual-inertial odometry approach for localization.  

\subsubsection{Camera}

\begin{figure}
    \centering
    \includegraphics[width=0.75\textwidth]{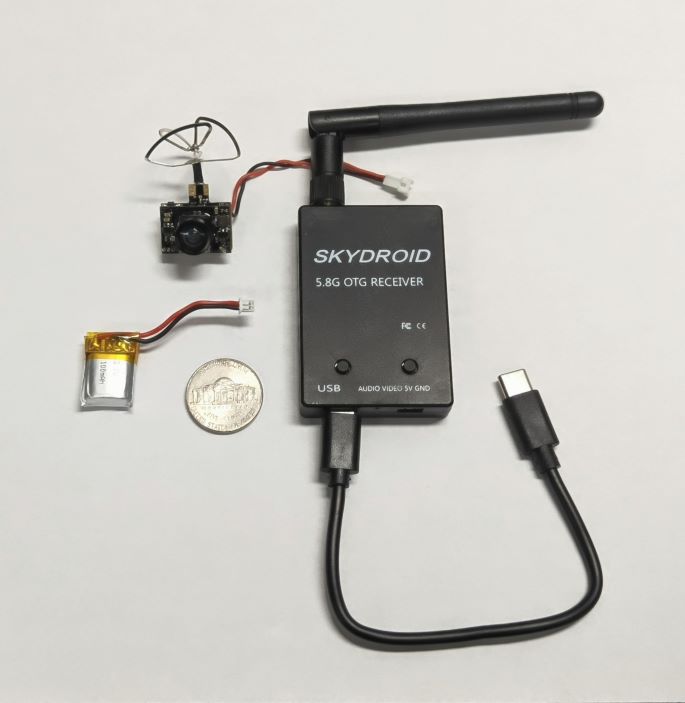}
    \caption{FPV camera used in early testing}
    \label{fig:fpvcamcomponents}
\end{figure}

\begin{figure}
    \centering
    \begin{subfigure}[b]{0.45\textwidth}
        \includegraphics[width=\textwidth, keepaspectratio]{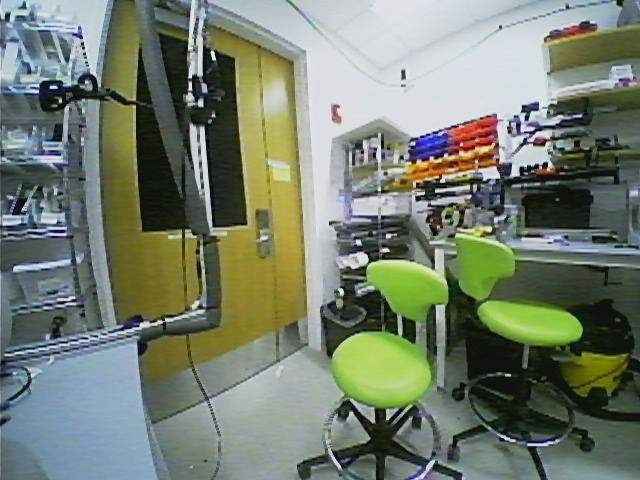}
        \caption{FPV Camera}
        \label{fig:fpvcampic}
    \end{subfigure}
    \hfill
    \begin{subfigure}[b]{0.45\textwidth}
        \includegraphics[width=\textwidth, keepaspectratio]{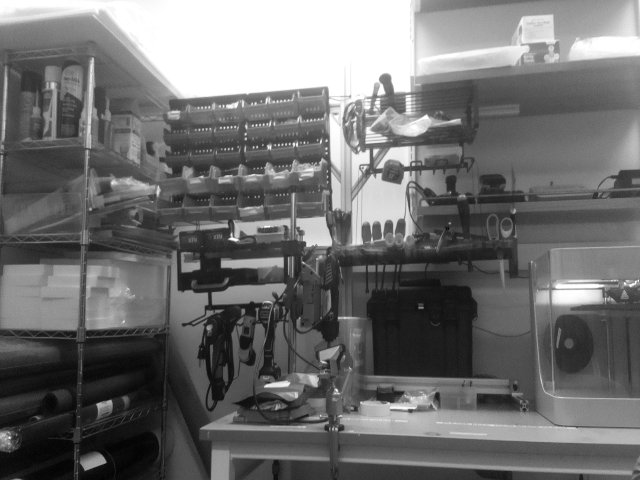}
        \caption{Raspberry Pi Camera}
        \label{fig:rpicampic}
    \end{subfigure}
    \caption{Pictures taken from about the same spot using FPV camera and Raspberry Pi Camera both with resolution 640x480}
    \label{fig:camera_pics_comparison}
\end{figure}

In the original configuration of Aerobat (Fig. \ref{fig:beta_old}), to get a sense of what images from an onboard camera would look like, a small FPV wireless camera was used (Fig. \ref{fig:fpvcamcomponents}). It consisted of the camera itself with an attached antenna and dedicated battery mounted on Aerobat, and a radio receiver connected to a laptop offboard through USB. This streamed 640x480 resolution image at 30Hz with no noticeable lag with line of sight communication. The camera weighs 4.53g. Fig. \ref{fig:fpvcampic} shows an image from this camera. 

Without a microprocessor onboard, this camera allowed us to see the world from Aerobat's perspective while it was flying. However, this could not be a long-term solution as the only interface to this camera was through the wireless receiver. Although a few options for camera modules were compared, with the Raspberry Pi Zero 2 W selected as the onboard processor, the natural choice was to use the Raspberry Pi Camera (Fig. \ref{fig:rpicamera}) as the interfaces were already in place. 

The Raspberry Pi Camera Model 2 weighs 3g and has an 8-megapixel sensor that offers video streaming up to 1080p at 30 fps and 720p at 60 fps. Fig. \ref{fig:camera_pics_comparison} compares images from the Raspberry Pi camera and the previously used FPV camera. While the FPV camera has a wider field of view, definition in features is lost in the center of the scene when compared with the Raspberry Pi Camera. The Raspberry Pi Camera also offers a few additional perks. It allows for setting the internal sensor update rate, independent of the rate at which images are read from the camera. This allows for low motion blur in images even when reading from the camera at low frame rates. It also has an internal GPU that gives it the ability to adjust exposure, shutter speed, brightness, contrast, saturation, and rotation, taking the load off the processor. Section \ref{sec:camera_integration} describes development of camera drivers.

\subsubsection{Inertial Measurement Unit (IMU)}

In this early stage of development, flight tests typically last just a few seconds at a time, removing IMU drift as a factor. However, there were two requirements for the IMU to meet: 

\begin{enumerate}
    \item \textbf{Data rate:} For good visual inertial odometry, it is ideal to have visual data at at around 20 Hz and inertial data at around 200 Hz. Therefore, the IMU must be able to provide data at 200 Hz or more.
    \item \textbf{Weight:} With 14g of payload taken up by the processor (11g) and camera (3g), there is only 1g of the imposed payload limit left for the IMU. Therefore, the IMU must weigh 1g or under. 
\end{enumerate}

Professional grade IMUs such as the VN-100 would be superfluous and expensive options at this stage when hobby-grade components are able to meet the requirements while being lighter in weight and lower in cost. Popular hobby-grade IMUs such as Adafruit's MPU6050 and ICM-20948 can comfortably meet these requirements, weighing just 1g and giving high data rates up to 400 kHz through I2C, limited only by the read/write speed of the interfaced processor. At the time of development, however, these were unavailable due to the ongoing chip shortage. Some IMUs such as Adafruit's BNO055 and WIT-motion's WT901 have similar weights and perform sensor fusion onboard to provide useful data such as absolute orientation, gravity-corrected linear acceleration, and gravity vectors. BNO055 only has a maximum rate of 100 Hz but WT901 has a maximum update rate of 200 Hz, meeting the desired data rate. Due to its availability and suitability to the requirements, this was chosen as the IMU for Aerobat (Fig. \ref{fig:imu}).

The sensor is interfaced by I2C communication and internally updates registers containing the following information:

\begin{itemize}
    \item Linear Acceleration (x, y, z)
    \item Angular Velocity (x, y, z)
    \item Magnetic Field Strength (x, y, z)
    \item Kalman Filtered Absolute Orientation in Euler angles (Roll, Pitch, Yaw)
    \item Kalman Filtered Absolute Orientation in Quaternion (x, y, z, w)
    \item Temperature
\end{itemize}

Each value is stored in 2 Bytes of memory, bringing a total of 34 Bytes of information available to be read in each sampling of the IMU. Section \ref{sec:imu_integration} describes how this data is read onboard by the IMU driver.

\begin{figure}
    \centering
    \includegraphics[width=\textwidth, keepaspectratio]{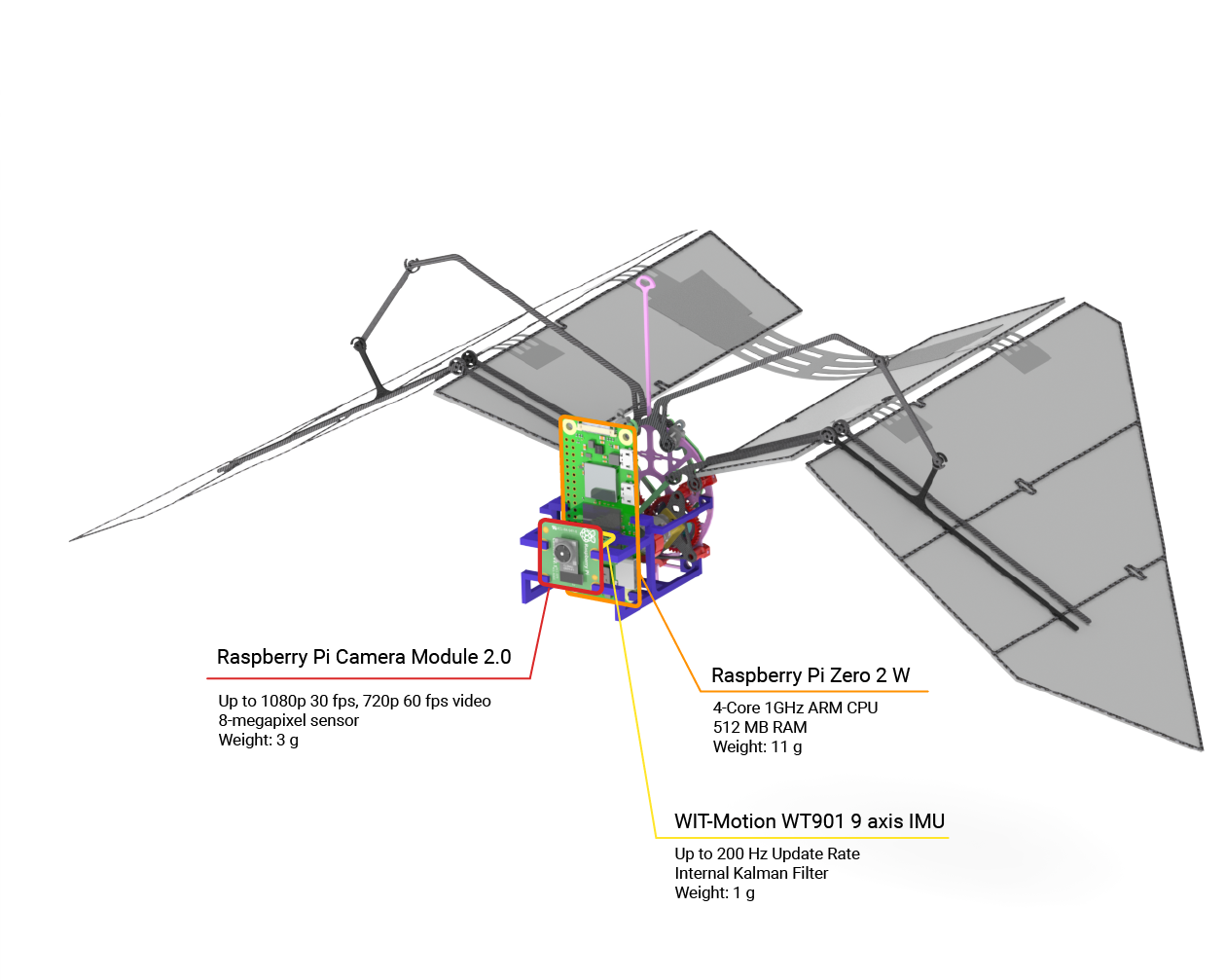}
    \caption{Model of Onboard Electronics on Aerobat}
    \label{fig:onboard_electronics}
\end{figure}

\begin{table}[]
    \centering
    \begin{tabular}{l c c}
        Aerobat Structure + Wings  &  22g\\
        Motor + ESC + Gearbox  & 8g\\
        Electronics Mount & 3g\\
        Battery & 18g\\
        Processor & 11g\\
        Camera & 3g\\
        IMU & 1g\\
        \textbf{Total} & \textbf{66g}
    \end{tabular}
    \caption{Breakdown of each of the components on Aerobat by weight}
    \label{tab:weight_breakdown}
\end{table}

\section{Sensor Integration}
\label{sec:sensor_integration}

Results from \cite{delmerico_benchmark_2018} indicated that most standard perception approaches would struggle to run on the Raspberry Pi Zero's 512 MB of RAM, and so initial efforts were focused on testing the capabilities of the processor, the camera, and the IMU.

\subsection{Camera}
\label{sec:camera_integration}

\begin{figure}
    \centering
    \includegraphics[width=0.75\textwidth]{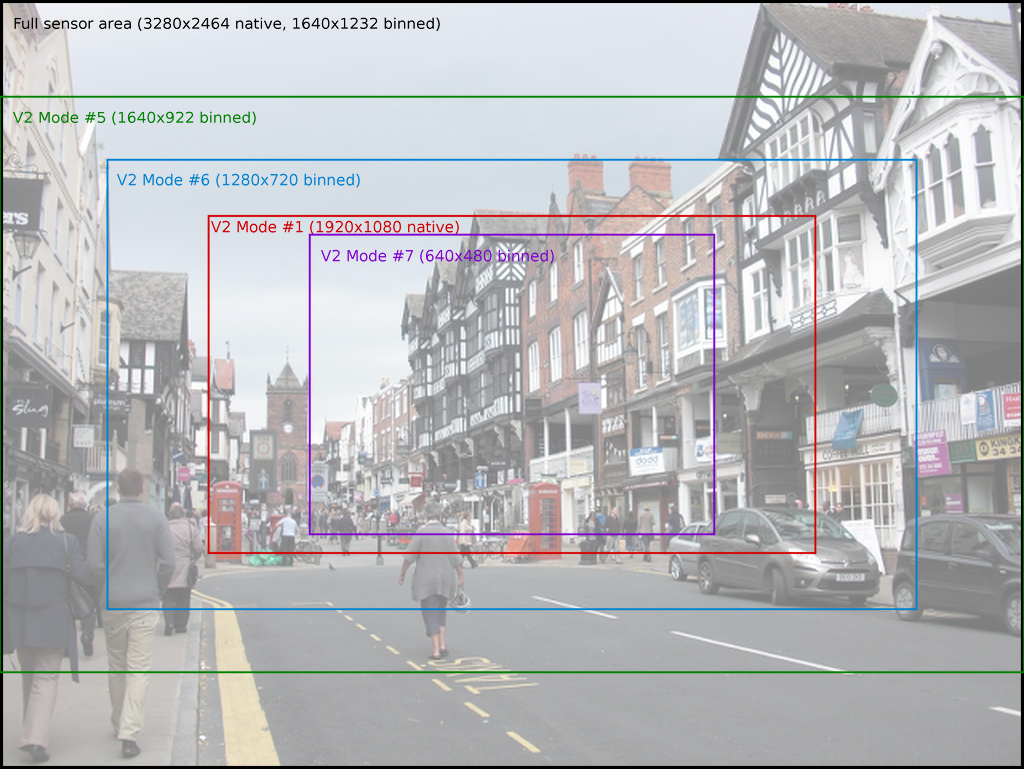}
    \caption{Depiction of active sensor areas at different resolutions from picamera documentation \cite{noauthor_6_nodate}}
    \label{fig:sensor_areas}
\end{figure}

Camera drivers were implemented onboard using Python3 and various algorithms were tested for performance with greyscale images at resolutions of 640x480 and 320x240 and 30 frames per second, including SIFT feature detection, Sparse and Dense Optical Flow, and Apriltag Detection. All of these comfortably ran onboard, utilizing less than 20\% of available memory. The field of view in these cases, however, appeared to be smaller than the field of view of the camera. Upon investigation, it was determined that the field of view was intentionally being clipped based on the resolution of the image and the framerate. This behaviour is described by the chart in Fig. \ref{fig:sensor_areas} from the Picamera documentation. To get around the clipped field of view issue, the driver was modified to read images at the full resolution (1640x922) and then resize it on the processor to the desired resolution. This gives the full field of view shown in Fig. \ref{fig:rpicampic}.

\subsection{IMU}
\label{sec:imu_integration}

IMU drivers were also implemented and interfaced using I2C communication and read rates of up to 1 kHz were achieved. Confident that the processor was capable of handling more, a bare-bones version of ROS Noetic was installed from source, containing only libraries required for the ROS perception stack.

\begin{figure}
    \centering
    \begin{subfigure}[b]{0.45\textwidth}
        \includegraphics[width=\textwidth, keepaspectratio]{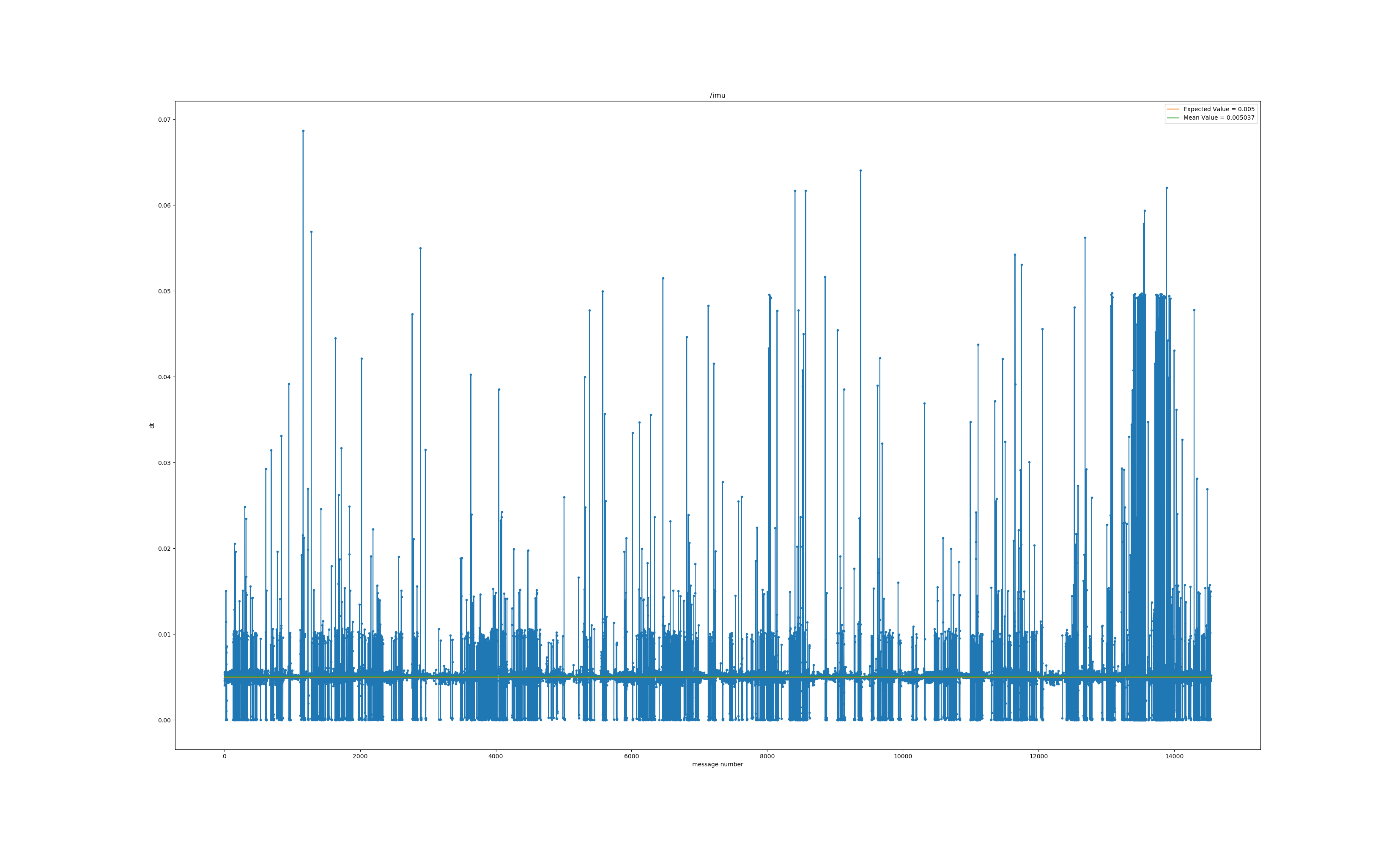}
        \caption{Only IMU node Running}
        \label{fig:200hzIMU}
    \end{subfigure}
    \hfill
    \begin{subfigure}[b]{0.45\textwidth}
        \includegraphics[width=\textwidth, keepaspectratio]{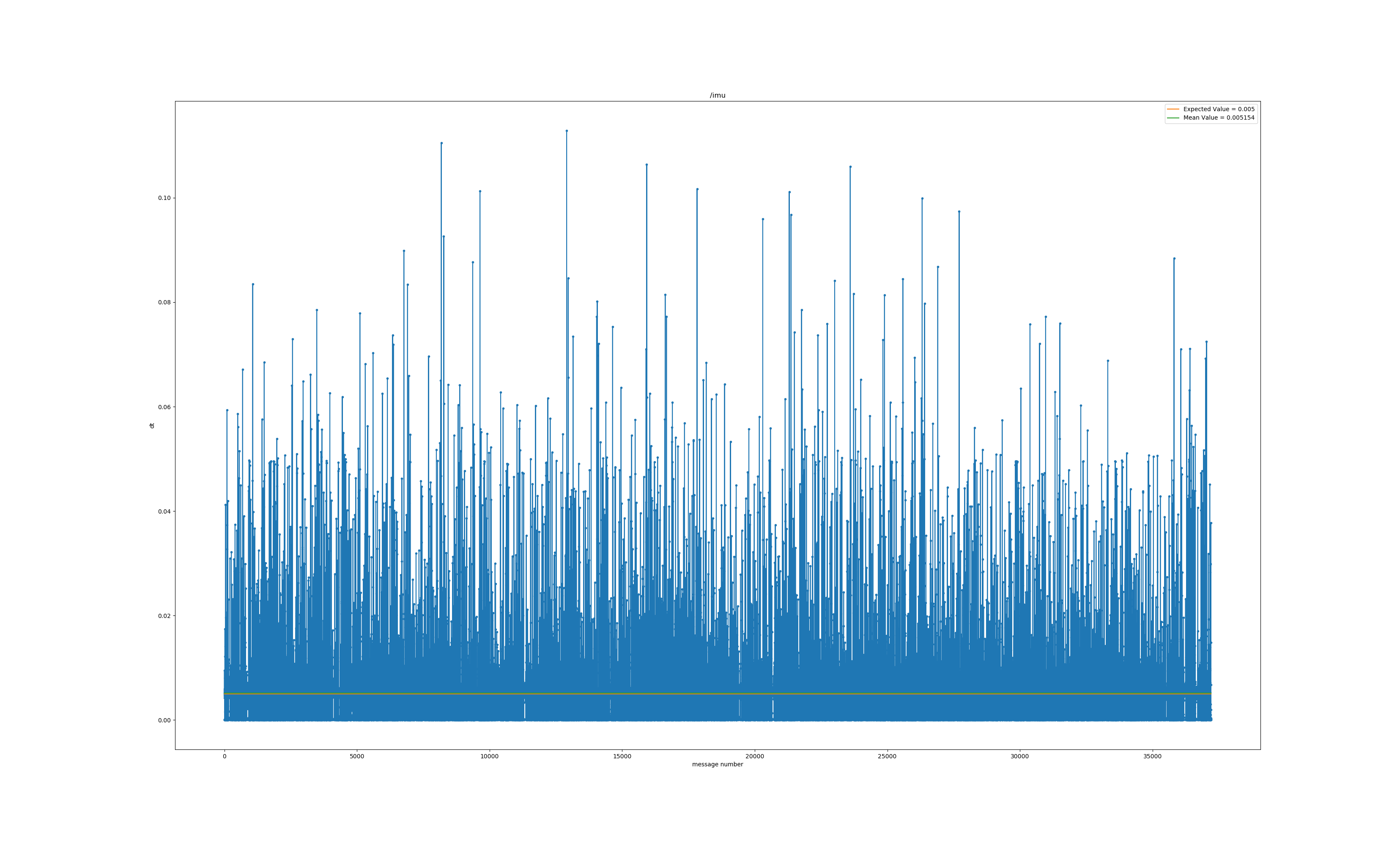}
        \caption{IMU and Camera node running}
        \label{fig:200hzIMUCamera}
    \end{subfigure}
    \caption{Plot of the time difference between every pair of consecutive data points in IMU data collected over 2 min at a rate of 200 Hz. The orange plot represents the expected value of 1/200 sec. Under ideal conditions, every data point would lie on this line.}
    \label{fig:timestamp_variation}
\end{figure}

Continuing further testing of the processor, IMU and camera ROS drivers were implemented and tested at various publishing rates. Initial results showed that while memory usage was acceptable at around 25\%, processing times were highly variable at higher rates. Figure \ref{fig:200hzIMU} shows the large variation in the time period of the IMU data being published over a period of 2 min of recording at a desired rate of 200 Hz. The variation in timestamps looks asymmetrical because of the way ROS handles timing. When a message takes longer than the desired rate to publish, the next message is published almost immediately with no delay. This leads to a number of messages with close to zero time difference from the previous message. This variation is further increased as the additional node to publish camera data is run alongside it \ref{fig:200hzIMUCamera}.

Investigation showed one of the causes to be long read times for I2C communication with the IMU. This was initially implemented as individual register reads for each of the data provided by the IMU, but optimization by reading a large contiguous block of data instead allowed reducing the number of I2C reads from 16 to 2. This considerably sped up the operation, improving performance.

\begin{figure}
    \centering
    \includegraphics[width=\textwidth, keepaspectratio]{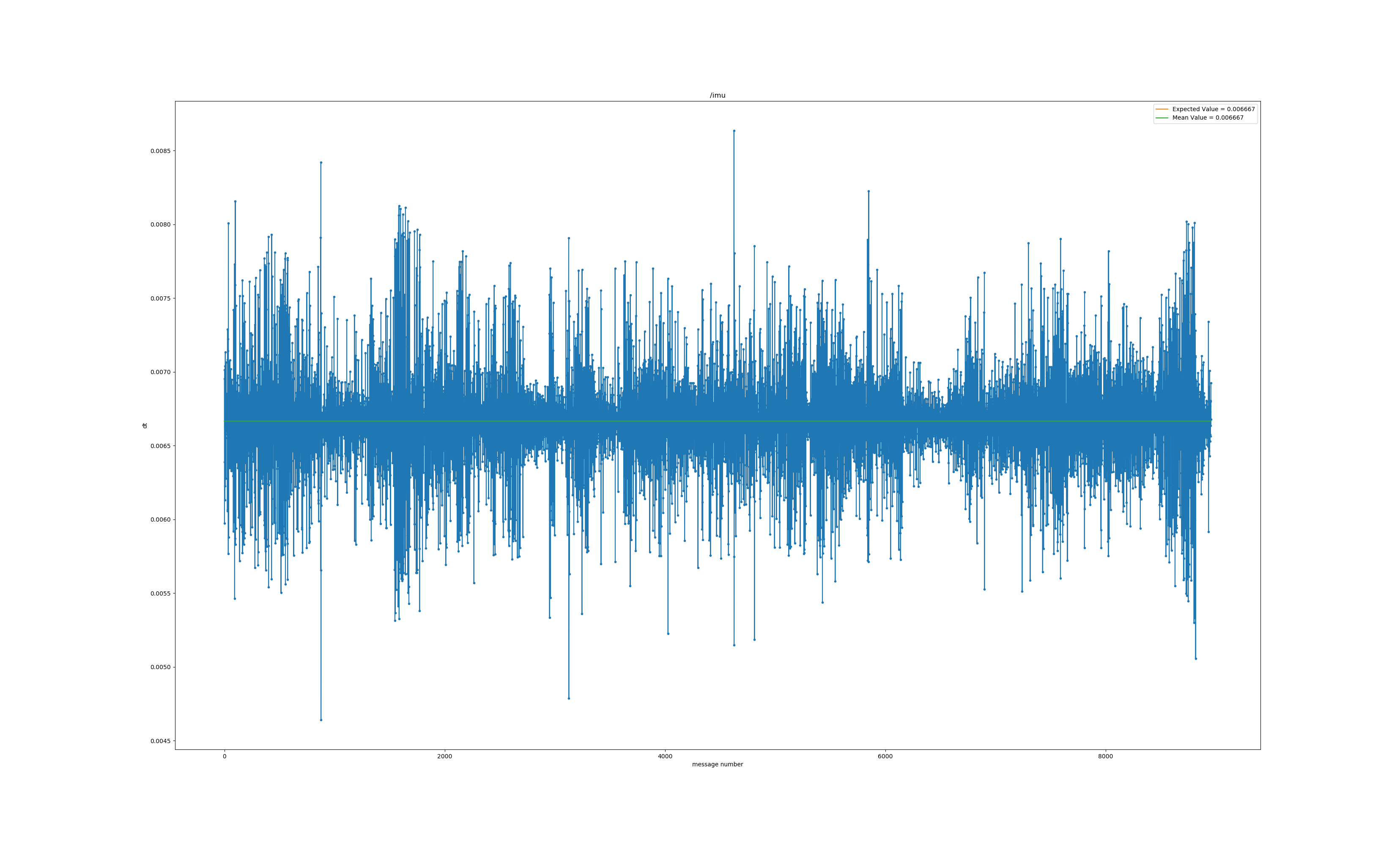}
    \caption{Timestamp variation in the IMU data at 150 Hz with the improved driver}
    \label{fig:newIMU150}
\end{figure}

Further improvements to the timing were made by using timed callback functions using ros::Timers rather than rate-based sleep functions in loops to read data from the sensors and reducing the frequency from 200 Hz to 150 Hz. Figure \ref{fig:newIMU150} shows the improved performance of the IMU at 150 Hz. 

There are still non-trivial variations in the time periods in the data. This is likely due to the number of parallel processes in operation. Running two nodes (one for camera and one for IMU) through ROS spawns over 10 threads, which, on a 4-core processor such as the Raspberry Pi Zero, would cause interruptions that increase the time between successive data reads. On a faster processor, this may not pose a challenge, with the processor able to keep up with the desired rate despite interruptions. However, this is likely a hardware limitation of the Raspberry Pi Zero.

\section{Sensor Calibration}

Camera and IMU were first individually calibrated. The standard checkerboard and Matlab's camera calibration toolbox was used to calibrate the camera and 6 hours of stationary bias-corrected data was used with \cite{noauthor_allan_2022} to calibrate the IMU.

With this calibration data, camera and IMU were calibrated together using Kalibr's \cite{noauthor_ethz-aslkalibr_nodate} camera-imu calibration script. Using a 4x6 1cm tag size aprilgrid as the calibration target, RISE Arena's manipulator was used to move the robot around, exciting all axes of the IMU. However, despite low camera re-projection errors and estimated accelerometer and gyroscope errors in the prior, optimization fails to find a solution for this setup. This is as yet an open issue, with potential sources of error being the same timing issues still affecting the data, IMU axes not being excited enough or the camera not getting a wide enough field of view for the data.

\section{Concluding remarks}

This chapter described the challenges in selecting and integrating electronics on a tight payload budget, and the challenges associated with running the perception stack onboard with limited hardware. With a fully integrated perception system, future work will be focused on utilizing RISE Arena (Section \ref{sec:rise}) to test VIO algorithms onboard and evaluating their feasibility and challenges in implementing autonomous flight for Aerobat. Work will also be required to integrate the kinematics and dynamics of Aerobat into the perception algorithm for more robust state estimation. Using these, Aerobat will be flown autonomously within RISE Arena using the perception system to stay within the boundaries of the confined space while performing aerial maneuvers.    


\chapter{Conclusion}
\label{chap:conclude}

This thesis presents the progress made towards Autonomous Untethered Flight on Northeastern University's Aerobat. This was broken down into three primary goals with the progress towards each described in their own chapter.

\section{Chapter \ref{chap:untethered}}
Chapter \ref{chap:untethered} described progress made towards untethered flight. A proof-of-concept 10m outdoor untethered flight was demonstrated and two additional development was presented, towards enabling future testing for untethered flight. The first of these was the protective guard, Kongming Lamp (Sec. \ref{sec:guard}), which was drop tested with a representative weight at the center to demonstrate protection for Aerobat from crashes. The second development was RISE Arena (Sec. \ref{sec:rise}), providing elaborate ground truth information for controlled and repeatable testing and system identification.

\subsection{Thesis Contributions}
For the outdoor untethered flight demonstration, stability of flight was improved by tuning the complementary filter applied to calculate orientation from IMU for stabilization. For the design of Kongming Lamp, in addition to conceptual inputs, control code for stabilization using IMU and pose information was developed and tuned. In addition, RISE Arena was fully developed as a part of this thesis, including interfacing and control code for the manipulator, calibration of Optitrack system and integration of processing and sensing onto Aerobat-Gamma.

\section{Chapter \ref{chap:control}}

Chapter \ref{chap:control} described the aerodynamic model of Aerobat and the steps taken towards validating the model. Preliminary results indicate the model is accurate, but further system identification is required to fully map out the control system and experimentally test the model in-flight under different wind conditions. Predicated on the success of this, outdoor closed-loop tests may be performed with the help of Kongming Lamp (Sec. \ref{sec:guard}) until Aerobat is ready to fly completely unsupported. 

\subsection{Thesis Contributions}
Created the manipulator trajectories and performed the experiment using RISE Arena to collect data for validation of the aerodynamic model.

\section{Chapter \ref{chap:perception}}

Chapter \ref{chap:perception} described the progress made towards onboard perception and state estimation. Processors and sensors were selected and integrated onto the robot (Sec. \ref{sec:electronics}). Sensor drivers were written and iteratively optimized for timing issues and speed of processing (Sec. \ref{sec:sensor_integration}). ROS was installed and tested on the limited processing power available on Aerobat and preliminary data for VIO was collected with the help of RISE Arena. As an immediate next goal, this data will be run through different VIO algorithms to verify the quality of the data and benchmark the algorithms.

\subsection{Thesis Contributions}
This chapter describes research and development fully carried out as part of this thesis.

\section{Future Work}
This work will be continued as part of my doctoral study, and as further progress is made on each of these goals, Aerobat will be at a mature stage where technology demonstrations may be made such as:
\begin{itemize}
    \item Controlled near ground flight akin to birds and bats demonstrating higher efficiency of near ground flight
    \item Long distance flights demonstrating the high efficiency of flapping wing systems
    \item Autonomous flight within a straight tunnel demonstrating precise closed loop control in confined areas
    \item Autonomous flight within a tunnel maze demonstrating the high agility of flapping wing systems and their ability to open up previously inaccessible spaces
\end{itemize}

\nocite{ramezani_bat_2016, ramezani_nonlinear_nodate, ramezani_modeling_2016, hoff_synergistic_2016, hoff_reducing_2017, ramezani_describing_2017, ramezani_biomimetic_2017, syed_rousettus_2017-1, hoff_optimizing_2018, hoff_trajectory_2019, ramezani_towards_2020}
\nocite{sihite_mechanism_2020, sihite_computational_2020, sihite_enforcing_2020, sihite_orientation_2021, ramezani_aerobat_2022, sihite_unsteady_2022}
\printbibliography

@misc{ramezani_aerobat_2022,
	title = {Aerobat, {A} {Bioinspired} {Drone} to {Test} {High}-{DOF} {Actuation} and {Embodied} {Aerial} {Locomotion}},
	url = {http://arxiv.org/abs/2212.05361},
	doi = {10.48550/arXiv.2212.05361},
	urldate = {2023-05-17},
	publisher = {arXiv},
	author = {Ramezani, Alireza and Sihite, Eric},
	month = dec,
	year = {2022},
	note = {arXiv:2212.05361 [cs, eess]},
}

@inproceedings{sihite_unsteady_2022,
	title = {Unsteady aerodynamic modeling of {Aerobat} using lifting line theory and {Wagner}'s function},
	doi = {10.1109/IROS47612.2022.9982125},
	booktitle = {2022 {IEEE}/{RSJ} {International} {Conference} on {Intelligent} {Robots} and {Systems} ({IROS})},
	author = {Sihite, Eric and Ghanem, Paul and Salagame, Adarsh and Ramezani, Alireza},
	month = oct,
	year = {2022},
	note = {ISSN: 2153-0866},
	pages = {10493--10500},
}

@misc{sihite_bang-bang_2022,
	title = {Bang-{Bang} {Control} {Of} {A} {Tail}-less {Morphing} {Wing} {Flight}},
	url = {http://arxiv.org/abs/2205.06395},
	doi = {10.48550/arXiv.2205.06395},
	urldate = {2023-05-17},
	publisher = {arXiv},
	author = {Sihite, Eric and Hu, Xintao and Li, Bozhen and Salagame, Adarsh and Ghanem, Paul and Ramezani, Alireza},
	month = may,
	year = {2022},
	note = {arXiv:2205.06395 [cs, eess]},
}

@misc{ghanem_efficient_2021,
	title = {Efficient {Modeling} of {Morphing} {Wing} {Flight} {Using} {Neural} {Networks} and {Cubature} {Rules}},
	url = {http://arxiv.org/abs/2110.01057},
	doi = {10.48550/arXiv.2110.01057},
	urldate = {2023-05-17},
	publisher = {arXiv},
	author = {Ghanem, Paul and Bicer, Yunus and Erdogmus, Deniz and Ramezani, Alireza},
	month = oct,
	year = {2021},
	note = {arXiv:2110.01057 [cs, eess]},
}

@inproceedings{sihite_orientation_2021,
	title = {Orientation stabilization in a bioinspired bat-robot using integrated mechanical intelligence and control},
	volume = {11758},
	url = {https://www.spiedigitallibrary.org/conference-proceedings-of-spie/11758/1175805/Orientation-stabilization-in-a-bioinspired-bat-robot-using-integrated-mechanical/10.1117/12.2587894.full},
	doi = {10.1117/12.2587894},
	urldate = {2023-05-17},
	booktitle = {Unmanned {Systems} {Technology} {XXIII}},
	publisher = {SPIE},
	author = {Sihite, Eric and Lessieur, Andrew and Dangol, Pravin and Singhal, Akshath and Ramezani, Alireza},
	month = apr,
	year = {2021},
	pages = {12--20},
}

@inproceedings{sihite_enforcing_2020,
	title = {Enforcing nonholonomic constraints in {Aerobat}, a roosting flapping wing model},
	doi = {10.1109/CDC42340.2020.9304158},
	booktitle = {2020 59th {IEEE} {Conference} on {Decision} and {Control} ({CDC})},
	author = {Sihite, Eric and Ramezani, Alireza},
	month = dec,
	year = {2020},
	note = {ISSN: 2576-2370},
	pages = {5321--5327},
}

@misc{sihite_mechanism_2020,
	title = {Mechanism {Design} of a {Bio}-inspired {Armwing} {Mechanism} for {Mimicking} {Bat} {Flapping} {Gait}},
	url = {http://arxiv.org/abs/2010.04702},
	doi = {10.48550/arXiv.2010.04702},
	urldate = {2023-05-17},
	publisher = {arXiv},
	author = {Sihite, E. and Kelly, P. and Ramezani, A.},
	month = oct,
	year = {2020},
	note = {arXiv:2010.04702 [cs]},
}

@article{sihite_computational_2020,
	title = {Computational {Structure} {Design} of a {Bio}-{Inspired} {Armwing} {Mechanism}},
	volume = {5},
	issn = {2377-3766},
	doi = {10.1109/LRA.2020.3010217},
	number = {4},
	journal = {IEEE Robotics and Automation Letters},
	author = {Sihite, Eric and Kelly, Peter and Ramezani, Alireza},
	month = oct,
	year = {2020},
	note = {Conference Name: IEEE Robotics and Automation Letters},
	pages = {5929--5936},
}

@inproceedings{ramezani_towards_2020,
	title = {Towards biomimicry of a bat-style perching maneuver on structures: the manipulation of inertial dynamics},
	shorttitle = {Towards biomimicry of a bat-style perching maneuver on structures},
	doi = {10.1109/ICRA40945.2020.9197376},
	booktitle = {2020 {IEEE} {International} {Conference} on {Robotics} and {Automation} ({ICRA})},
	author = {Ramezani, Alireza},
	month = may,
	year = {2020},
	note = {ISSN: 2577-087X},
	pages = {7015--7021},
}

@inproceedings{hoff_trajectory_2019,
	title = {Trajectory planning for a bat-like flapping wing robot},
	doi = {10.1109/IROS40897.2019.8968450},
	booktitle = {2019 {IEEE}/{RSJ} {International} {Conference} on {Intelligent} {Robots} and {Systems} ({IROS})},
	author = {Hoff, Jonathan and Syed, Usman and Ramezani, Alireza and Hutchinson, Seth},
	month = nov,
	year = {2019},
	note = {ISSN: 2153-0866},
	pages = {6800--6805},
}

@article{hoff_optimizing_2018,
	title = {Optimizing the structure and movement of a robotic bat with biological kinematic synergies},
	volume = {37},
	issn = {0278-3649},
	url = {https://doi.org/10.1177/0278364918804654},
	doi = {10.1177/0278364918804654},
	language = {en},
	number = {10},
	urldate = {2023-05-17},
	journal = {The International Journal of Robotics Research},
	author = {Hoff, Jonathan and Ramezani, Alireza and Chung, Soon-Jo and Hutchinson, Seth},
	month = sep,
	year = {2018},
	note = {Publisher: SAGE Publications Ltd STM},
	pages = {1233--1252},
}

@article{ramezani_biomimetic_2017,
	title = {A biomimetic robotic platform to study flight specializations of bats},
	volume = {2},
	url = {https://www.science.org/doi/full/10.1126/scirobotics.aal2505},
	doi = {10.1126/scirobotics.aal2505},
	number = {3},
	urldate = {2023-05-17},
	journal = {Science Robotics},
	author = {Ramezani, Alireza and Chung, Soon-Jo and Hutchinson, Seth},
	month = feb,
	year = {2017},
	note = {Publisher: American Association for the Advancement of Science},
	pages = {eaal2505},
}

@inproceedings{hoff_reducing_2017,
	address = {Cham},
	series = {Lecture {Notes} in {Computer} {Science}},
	title = {Reducing {Versatile} {Bat} {Wing} {Conformations} to a 1-{DoF} {Machine}},
	isbn = {978-3-319-63537-8},
	doi = {10.1007/978-3-319-63537-8_16},
	language = {en},
	booktitle = {Biomimetic and {Biohybrid} {Systems}},
	publisher = {Springer International Publishing},
	author = {Hoff, Jonathan and Ramezani, Alireza and Chung, Soon-Jo and Hutchinson, Seth},
	editor = {Mangan, Michael and Cutkosky, Mark and Mura, Anna and Verschure, Paul F.M.J. and Prescott, Tony and Lepora, Nathan},
	year = {2017},
	pages = {181--192},
}

@inproceedings{ramezani_describing_2017,
	address = {Cham},
	series = {Lecture {Notes} in {Computer} {Science}},
	title = {Describing {Robotic} {Bat} {Flight} with {Stable} {Periodic} {Orbits}},
	isbn = {978-3-319-63537-8},
	doi = {10.1007/978-3-319-63537-8_33},
	language = {en},
	booktitle = {Biomimetic and {Biohybrid} {Systems}},
	publisher = {Springer International Publishing},
	author = {Ramezani, Alireza and Ahmed, Syed Usman and Hoff, Jonathan and Chung, Soon-Jo and Hutchinson, Seth},
	editor = {Mangan, Michael and Cutkosky, Mark and Mura, Anna and Verschure, Paul F.M.J. and Prescott, Tony and Lepora, Nathan},
	year = {2017},
	pages = {394--405},
}

@inproceedings{hoff_synergistic_2016,
	title = {Synergistic {Design} of a {Bio}-{Inspired} {Micro} {Aerial} {Vehicle} with {Articulated} {Wings}},
	isbn = {978-0-9923747-2-3},
	url = {http://www.roboticsproceedings.org/rss12/p09.pdf},
	doi = {10.15607/RSS.2016.XII.009},
	language = {en},
	urldate = {2023-05-17},
	booktitle = {Robotics: {Science} and {Systems} {XII}},
	publisher = {Robotics: Science and Systems Foundation},
	author = {Hoff, Jonathan and Ramezani, Alireza and Chung, Soon-Jo and Hutchinson, Seth},
	year = {2016},
}

@inproceedings{ramezani_bat_2016,
	title = {Bat {Bot} ({B2}), a biologically inspired flying machine},
	doi = {10.1109/ICRA.2016.7487491},
	booktitle = {2016 {IEEE} {International} {Conference} on {Robotics} and {Automation} ({ICRA})},
	author = {Ramezani, Alireza and Shi, Xichen and Chung, Soon-Jo and Hutchinson, Seth},
	month = may,
	year = {2016},
	pages = {3219--3226},
}

@incollection{ramezani_nonlinear_nodate,
	title = {Nonlinear {Flight} {Controller} {Synthesis} of a {Bat}-{Inspired} {Micro} {Aerial} {Vehicle}},
	url = {https://arc.aiaa.org/doi/abs/10.2514/6.2016-1376},
	urldate = {2023-05-17},
	booktitle = {{AIAA} {Guidance}, {Navigation}, and {Control} {Conference}},
	publisher = {American Institute of Aeronautics and Astronautics},
	author = {Ramezani, Alireza and Shi, Xichen and Chung, Soon-Jo and Hutchinson, Seth},
	doi = {10.2514/6.2016-1376},
	note = {\_eprint: https://arc.aiaa.org/doi/pdf/10.2514/6.2016-1376},
}

@inproceedings{ramezani_modeling_2016,
	title = {Modeling and nonlinear flight controller synthesis of a bat-inspired micro aerial vehicle},
	url = {https://experts.illinois.edu/en/publications/modeling-and-nonlinear-flight-controller-synthesis-of-a-bat-inspi},
	language = {English (US)},
	urldate = {2020-05-09},
	booktitle = {{AIAA} {Guidance}, {Navigation}, and {Control} {Conference}},
	publisher = {American Institute of Aeronautics and Astronautics Inc, AIAA},
	author = {Ramezani, Alireza and Shi, Xichen and Chung, Soon-Jo and Hutchinson, Seth Andrew},
	month = jan,
	year = {2016},
}

@inproceedings{syed_rousettus_2017-1,
	address = {Singapore, Singapore},
	title = {From {Rousettus} aegyptiacus (bat) landing to robotic landing: {Regulation} of {CG}-{CP} distance using a nonlinear closed-loop feedback},
	isbn = {978-1-5090-4633-1},
	shorttitle = {From {Rousettus} aegyptiacus (bat) landing to robotic landing},
	url = {http://ieeexplore.ieee.org/document/7989408/},
	doi = {10.1109/ICRA.2017.7989408},
	language = {en},
	urldate = {2020-01-04},
	booktitle = {2017 {IEEE} {International} {Conference} on {Robotics} and {Automation} ({ICRA})},
	publisher = {IEEE},
	author = {Syed, Usman A. and Ramezani, Alireza and Chung, Soon-Jo and Hutchinson, Seth},
	month = may,
	year = {2017},
	pages = {3560--3567},
}

@misc{noauthor_6_nodate,
	title = {6. {Camera} {Hardware} — {Picamera} 1.13 {Documentation}},
	url = {https://picamera.readthedocs.io/en/latest/fov.html#sensor-modes},
	urldate = {2022-08-18},
}

@article{de_croon_flapping_2020,
	title = {Flapping wing drones show off their skills},
	volume = {5},
	url = {https://www.science.org/doi/full/10.1126/scirobotics.abd0233},
	doi = {10.1126/scirobotics.abd0233},
	number = {44},
	urldate = {2022-08-17},
	journal = {Science Robotics},
	author = {de Croon, Guido},
	month = jul,
	year = {2020},
	note = {Publisher: American Association for the Advancement of Science},
	pages = {eabd0233},
}

@article{matus-vargas_ground_2021,
	title = {Ground effect on rotorcraft unmanned aerial vehicles: a review},
	volume = {14},
	issn = {1861-2784},
	shorttitle = {Ground effect on rotorcraft unmanned aerial vehicles},
	url = {https://doi.org/10.1007/s11370-020-00344-5},
	doi = {10.1007/s11370-020-00344-5},
	language = {en},
	number = {1},
	urldate = {2022-08-17},
	journal = {Intelligent Service Robotics},
	author = {Matus-Vargas, Antonio and Rodriguez-Gomez, Gustavo and Martinez-Carranza, Jose},
	month = mar,
	year = {2021},
	pages = {99--118},
}

@misc{noauthor_ethz-aslkalibr_nodate,
	title = {ethz-asl/kalibr: {The} {Kalibr} visual-inertial calibration toolbox},
	url = {https://github.com/ethz-asl/kalibr},
	urldate = {2022-08-15},
}

@misc{noauthor_allan_2022,
	title = {Allan {Variance} {ROS}},
	copyright = {BSD-3-Clause},
	url = {https://github.com/ori-drs/allan_variance_ros},
	urldate = {2022-08-15},
	publisher = {Oxford Dynamic Robot Systems Group},
	month = aug,
	year = {2022},
	note = {original-date: 2021-11-12T15:11:18Z},
}

@article{boutet_unsteady_2018,
	title = {Unsteady {Lifting} {Line} {Theory} {Using} the {Wagner} {Function} for the {Aerodynamic} and {Aeroelastic} {Modeling} of {3D} {Wings}},
	volume = {5},
	copyright = {http://creativecommons.org/licenses/by/3.0/},
	issn = {2226-4310},
	url = {https://www.mdpi.com/2226-4310/5/3/92},
	doi = {10.3390/aerospace5030092},
	language = {en},
	number = {3},
	urldate = {2022-08-15},
	journal = {Aerospace},
	author = {Boutet, Johan and Dimitriadis, Grigorios},
	month = sep,
	year = {2018},
	note = {Number: 3
Publisher: Multidisciplinary Digital Publishing Institute},
	pages = {92},
}

@article{kaess_isam2_2012,
	title = {{iSAM2}: {Incremental} smoothing and mapping using the {Bayes} tree},
	volume = {31},
	issn = {0278-3649},
	shorttitle = {{iSAM2}},
	url = {https://doi.org/10.1177/0278364911430419},
	doi = {10.1177/0278364911430419},
	language = {en},
	number = {2},
	urldate = {2022-08-15},
	journal = {The International Journal of Robotics Research},
	author = {Kaess, Michael and Johannsson, Hordur and Roberts, Richard and Ila, Viorela and Leonard, John J and Dellaert, Frank},
	month = feb,
	year = {2012},
	note = {Publisher: SAGE Publications Ltd STM},
	pages = {216--235},
}

@article{bloesch_iterated_2017,
	title = {Iterated extended {Kalman} filter based visual-inertial odometry using direct photometric feedback},
	volume = {36},
	issn = {0278-3649},
	url = {https://doi.org/10.1177/0278364917728574},
	doi = {10.1177/0278364917728574},
	language = {en},
	number = {10},
	urldate = {2022-08-15},
	journal = {The International Journal of Robotics Research},
	author = {Bloesch, Michael and Burri, Michael and Omari, Sammy and Hutter, Marco and Siegwart, Roland},
	month = sep,
	year = {2017},
	note = {Publisher: SAGE Publications Ltd STM},
	pages = {1053--1072},
}

@article{weinstein_visual_2018,
	title = {Visual {Inertial} {Odometry} {Swarm}: {An} {Autonomous} {Swarm} of {Vision}-{Based} {Quadrotors}},
	volume = {3},
	issn = {2377-3766},
	shorttitle = {Visual {Inertial} {Odometry} {Swarm}},
	doi = {10.1109/LRA.2018.2800119},
	number = {3},
	journal = {IEEE Robotics and Automation Letters},
	author = {Weinstein, Aaron and Cho, Adam and Loianno, Giuseppe and Kumar, Vijay},
	month = jul,
	year = {2018},
	note = {Conference Name: IEEE Robotics and Automation Letters},
	pages = {1801--1807},
}

@article{rhodes_autonomous_2022,
	title = {Autonomous {Source} {Term} {Estimation} in {Unknown} {Environments}: {From} a {Dual} {Control} {Concept} to {UAV} {Deployment}},
	volume = {7},
	issn = {2377-3766},
	shorttitle = {Autonomous {Source} {Term} {Estimation} in {Unknown} {Environments}},
	doi = {10.1109/LRA.2022.3143890},
	number = {2},
	journal = {IEEE Robotics and Automation Letters},
	author = {Rhodes, Callum and Liu, Cunjia and Chen, Wen-Hua},
	month = apr,
	year = {2022},
	note = {Conference Name: IEEE Robotics and Automation Letters},
	pages = {2274--2281},
}

@article{di_luca_bioinspired_2017,
	title = {Bioinspired morphing wings for extended flight envelope and roll control of small drones},
	volume = {7},
	url = {https://royalsocietypublishing.org/doi/full/10.1098/rsfs.2016.0092},
	doi = {10.1098/rsfs.2016.0092},
	number = {1},
	urldate = {2022-08-14},
	journal = {Interface Focus},
	author = {Di Luca, M. and Mintchev, S. and Heitz, G. and Noca, F. and Floreano, D.},
	month = feb,
	year = {2017},
	note = {Publisher: Royal Society},
	pages = {20160092},
}

@article{tedrake_learning_nodate,
	title = {Learning to {Fly} like a {Bird}},
	language = {en},
	author = {Tedrake, Russ and Jackowski, Zack and Cory, Rick and Roberts, John William and Hoburg, Warren},
	pages = {8},
}

@article{chang_soft_2020,
	title = {Soft biohybrid morphing wings with feathers underactuated by wrist and finger motion},
	volume = {5},
	url = {https://www.science.org/doi/full/10.1126/scirobotics.aay1246},
	doi = {10.1126/scirobotics.aay1246},
	number = {38},
	urldate = {2022-08-14},
	journal = {Science Robotics},
	author = {Chang, Eric and Matloff, Laura Y. and Stowers, Amanda K. and Lentink, David},
	month = jan,
	year = {2020},
	note = {Publisher: American Association for the Advancement of Science},
	pages = {eaay1246},
}

@article{gerdes_robo_2014,
	title = {Robo {Raven}: {A} {Flapping}-{Wing} {Air} {Vehicle} with {Highly} {Compliant} and {Independently} {Controlled} {Wings}},
	volume = {1},
	issn = {2169-5172},
	shorttitle = {Robo {Raven}},
	url = {https://www.liebertpub.com/doi/full/10.1089/soro.2014.0019},
	doi = {10.1089/soro.2014.0019},
	number = {4},
	urldate = {2022-08-14},
	journal = {Soft Robotics},
	author = {Gerdes, John and Holness, Alex and Perez-Rosado, Ariel and Roberts, Luke and Greisinger, Adrian and Barnett, Eli and Kempny, Johannes and Lingam, Deepak and Yeh, Chen-Haur and Bruck, Hugh A. and Gupta, Satyandra K.},
	month = dec,
	year = {2014},
	note = {Publisher: Mary Ann Liebert, Inc., publishers},
	pages = {275--288},
}

@inproceedings{peterson_experimental_2011,
	title = {Experimental dynamics of wing assisted running for a bipedal ornithopter},
	doi = {10.1109/IROS.2011.6095041},
	booktitle = {2011 {IEEE}/{RSJ} {International} {Conference} on {Intelligent} {Robots} and {Systems}},
	author = {Peterson, Kevin and Fearing, Ronald S.},
	month = sep,
	year = {2011},
	note = {ISSN: 2153-0866},
	pages = {5080--5086},
}

@article{phan_kubeetle-s_2019,
	title = {{KUBeetle}-{S}: {An} insect-like, tailless, hover-capable robot that can fly with a low-torque control mechanism},
	volume = {11},
	issn = {1756-8293},
	shorttitle = {{KUBeetle}-{S}},
	url = {https://doi.org/10.1177/1756829319861371},
	doi = {10.1177/1756829319861371},
	language = {en},
	urldate = {2022-08-14},
	journal = {International Journal of Micro Air Vehicles},
	author = {Phan, Hoang Vu and Aurecianus, Steven and Kang, Taesam and Park, Hoon Cheol},
	month = jan,
	year = {2019},
	note = {Publisher: SAGE Publications Ltd STM},
	pages = {1756829319861371},
}

@misc{noauthor_insectothopter_nodate,
	title = {Insectothopter - {CIA}},
	url = {https://www.cia.gov/legacy/museum/artifact/insectothopter/},
	urldate = {2022-08-14},
}

@article{ristroph_stable_2014,
	title = {Stable hovering of a jellyfish-like flying machine},
	volume = {11},
	url = {https://royalsocietypublishing.org/doi/full/10.1098/rsif.2013.0992},
	doi = {10.1098/rsif.2013.0992},
	number = {92},
	urldate = {2022-08-14},
	journal = {Journal of The Royal Society Interface},
	author = {Ristroph, Leif and Childress, Stephen},
	month = mar,
	year = {2014},
	note = {Publisher: Royal Society},
	pages = {20130992},
}

@article{de_croon_design_2009,
	title = {Design, {Aerodynamics}, and {Vision}-{Based} {Control} of the {DelFly}},
	volume = {1},
	issn = {1756-8293},
	url = {https://doi.org/10.1260/175682909789498288},
	doi = {10.1260/175682909789498288},
	language = {en},
	number = {2},
	urldate = {2022-08-14},
	journal = {International Journal of Micro Air Vehicles},
	author = {de Croon, G.C.H.E. and de Clercq, K.M.E. and Ruijsink, R. and Remes, B. and de Wagter, C.},
	month = jun,
	year = {2009},
	note = {Publisher: SAGE Publications Ltd STM},
	pages = {71--97},
}

@inproceedings{rosen_development_2016,
	title = {Development of a 3.2g untethered flapping-wing platform for flight energetics and control experiments},
	doi = {10.1109/ICRA.2016.7487492},
	booktitle = {2016 {IEEE} {International} {Conference} on {Robotics} and {Automation} ({ICRA})},
	author = {Rosen, Michelle H. and le Pivain, Geoffroy and Sahai, Ranjana and Jafferis, Noah T. and Wood, Robert J.},
	month = may,
	year = {2016},
	pages = {3227--3233},
}

@article{foehn_time-optimal_2021,
	title = {Time-optimal planning for quadrotor waypoint flight},
	volume = {6},
	url = {https://www.science.org/doi/full/10.1126/scirobotics.abh1221},
	doi = {10.1126/scirobotics.abh1221},
	number = {56},
	urldate = {2022-08-06},
	journal = {Science Robotics},
	author = {Foehn, Philipp and Romero, Angel and Scaramuzza, Davide},
	month = jul,
	year = {2021},
	pages = {eabh1221},
}

@article{foehn_alphapilot_2022,
	title = {{AlphaPilot}: autonomous drone racing},
	volume = {46},
	issn = {1573-7527},
	shorttitle = {{AlphaPilot}},
	url = {https://doi.org/10.1007/s10514-021-10011-y},
	doi = {10.1007/s10514-021-10011-y},
	language = {en},
	number = {1},
	urldate = {2022-08-06},
	journal = {Autonomous Robots},
	author = {Foehn, Philipp and Brescianini, Dario and Kaufmann, Elia and Cieslewski, Titus and Gehrig, Mathias and Muglikar, Manasi and Scaramuzza, Davide},
	month = jan,
	year = {2022},
	pages = {307--320},
}

@article{best_resilient_2022,
	title = {Resilient {Multi}-{Sensor} {Exploration} of {Multifarious} {Environments} with a {Team} of {Aerial} {Robots}},
	copyright = {info:eu-repo/semantics/openAccess},
	url = {https://opus.lib.uts.edu.au/handle/10453/157741},
	language = {en},
	urldate = {2022-08-06},
	author = {Best, G. and Garg, R. and Keller, J. and Hollinger, G. A. and Scherer, S.},
	year = {2022},
}

@article{petrlik_robust_2020,
	title = {A {Robust} {UAV} {System} for {Operations} in a {Constrained} {Environment}},
	volume = {5},
	issn = {2377-3766},
	doi = {10.1109/LRA.2020.2970980},
	number = {2},
	journal = {IEEE Robotics and Automation Letters},
	author = {Petrlík, Matěj and Báča, Tomáš and Heřt, Daniel and Vrba, Matouš and Krajník, Tomáš and Saska, Martin},
	month = apr,
	year = {2020},
	pages = {2169--2176},
}

@article{chen_aerial_2022,
	title = {Aerial {Grasping} and the {Velocity} {Sufficiency} {Region}},
	volume = {7},
	issn = {2377-3766},
	doi = {10.1109/LRA.2022.3192652},
	number = {4},
	journal = {IEEE Robotics and Automation Letters},
	author = {Chen, Tony G. and Hoffmann, Kenneth A. W. and Low, Jun En and Nagami, Keiko and Lentink, David and Cutkosky, Mark R.},
	month = oct,
	year = {2022},
	pages = {10009--10016},
}

@inproceedings{kaufmann_beauty_2019,
	title = {Beauty and the {Beast}: {Optimal} {Methods} {Meet} {Learning} for {Drone} {Racing}},
	shorttitle = {Beauty and the {Beast}},
	doi = {10.1109/ICRA.2019.8793631},
	booktitle = {2019 {International} {Conference} on {Robotics} and {Automation} ({ICRA})},
	author = {Kaufmann, Elia and Gehrig, Mathias and Foehn, Philipp and Ranftl, René and Dosovitskiy, Alexey and Koltun, Vladlen and Scaramuzza, Davide},
	month = may,
	year = {2019},
	note = {ISSN: 2577-087X},
	pages = {690--696},
}

@article{foehn_agilicious_2022,
	title = {Agilicious: {Open}-source and open-hardware agile quadrotor for vision-based flight},
	volume = {7},
	shorttitle = {Agilicious},
	url = {https://www.science.org/doi/full/10.1126/scirobotics.abl6259},
	doi = {10.1126/scirobotics.abl6259},
	number = {67},
	urldate = {2022-08-06},
	journal = {Science Robotics},
	author = {Foehn, Philipp and Kaufmann, Elia and Romero, Angel and Penicka, Robert and Sun, Sihao and Bauersfeld, Leonard and Laengle, Thomas and Cioffi, Giovanni and Song, Yunlong and Loquercio, Antonio and Scaramuzza, Davide},
	month = jun,
	year = {2022},
	pages = {eabl6259},
}

@article{tang_aggressive_2018,
	title = {Aggressive {Flight} {With} {Suspended} {Payloads} {Using} {Vision}-{Based} {Control}},
	volume = {3},
	issn = {2377-3766},
	doi = {10.1109/LRA.2018.2793305},
	number = {2},
	journal = {IEEE Robotics and Automation Letters},
	author = {Tang, Sarah and Wüest, Valentin and Kumar, Vijay},
	month = apr,
	year = {2018},
	pages = {1152--1159},
}

@inproceedings{engel_lsd-slam_2014,
	address = {Cham},
	series = {Lecture {Notes} in {Computer} {Science}},
	title = {{LSD}-{SLAM}: {Large}-{Scale} {Direct} {Monocular} {SLAM}},
	isbn = {978-3-319-10605-2},
	shorttitle = {{LSD}-{SLAM}},
	doi = {10.1007/978-3-319-10605-2_54},
	language = {en},
	booktitle = {Computer {Vision} – {ECCV} 2014},
	publisher = {Springer International Publishing},
	author = {Engel, Jakob and Schöps, Thomas and Cremers, Daniel},
	editor = {Fleet, David and Pajdla, Tomas and Schiele, Bernt and Tuytelaars, Tinne},
	year = {2014},
	pages = {834--849},
}

@article{hu_bang-bang_2022,
	title = {Bang-bang control of a tail-less morphing wing flight},
	url = {https://repository.library.northeastern.edu/files/neu:4f17gn39n},
	urldate = {2022-08-01},
	author = {Hu, Xintao},
	year = {2022},
}

@article{qin_vins-mono_2018,
	title = {{VINS}-{Mono}: {A} {Robust} and {Versatile} {Monocular} {Visual}-{Inertial} {State} {Estimator}},
	volume = {34},
	issn = {15523098},
	doi = {10.1109/TRO.2018.2853729},
	number = {4},
	journal = {IEEE Transactions on Robotics},
	author = {Qin, Tong and Li, Peiliang and Shen, Shaojie},
	month = aug,
	year = {2018},
	note = {arXiv: 1708.03852
Publisher: Institute of Electrical and Electronics Engineers Inc.},
	pages = {1004--1020},
}

@article{campos_orb-slam3_2021,
	title = {{ORB}-{SLAM3}: {An} {Accurate} {Open}-{Source} {Library} for {Visual}, {Visual}-{Inertial} and {Multi}-{Map} {SLAM}},
	volume = {37},
	issn = {1552-3098, 1941-0468},
	shorttitle = {{ORB}-{SLAM3}},
	url = {http://arxiv.org/abs/2007.11898},
	doi = {10.1109/TRO.2021.3075644},
	number = {6},
	urldate = {2022-07-17},
	journal = {IEEE Transactions on Robotics},
	author = {Campos, Carlos and Elvira, Richard and Rodríguez, Juan J. Gómez and Montiel, José M. M. and Tardós, Juan D.},
	month = dec,
	year = {2021},
	note = {arXiv:2007.11898 [cs]},
	pages = {1874--1890},
}

@article{zufferey_design_2021,
	title = {Design of the {High}-{Payload} {Flapping} {Wing} {Robot} {E}-{Flap}},
	volume = {6},
	issn = {2377-3766},
	doi = {10.1109/LRA.2021.3061373},
	number = {2},
	journal = {IEEE Robotics and Automation Letters},
	author = {Zufferey, Raphael and Tormo-Barbero, Jesús and Guzmán, M. Mar and Maldonado, Fco. Javier and Sanchez-Laulhe, Ernesto and Grau, Pedro and Pérez, Martín and Acosta, José Ángel and Ollero, Anibal},
	month = apr,
	year = {2021},
	note = {Conference Name: IEEE Robotics and Automation Letters},
	pages = {3097--3104},
}

@inproceedings{duhamel_altitude_2012,
	title = {Altitude feedback control of a flapping-wing microrobot using an on-board biologically inspired optical flow sensor},
	doi = {10.1109/ICRA.2012.6225313},
	booktitle = {2012 {IEEE} {International} {Conference} on {Robotics} and {Automation}},
	author = {Duhamel, Pierre-Emile J. and Pérez-Arancibia, Néstor O. and Barrows, Geoffrey L. and Wood, Robert J.},
	month = may,
	year = {2012},
	note = {ISSN: 1050-4729},
	pages = {4228--4235},
}

@inproceedings{garcia_bermudez_optical_2009,
	title = {Optical flow on a flapping wing robot},
	doi = {10.1109/IROS.2009.5354337},
	booktitle = {2009 {IEEE}/{RSJ} {International} {Conference} on {Intelligent} {Robots} and {Systems}},
	author = {Garcia Bermudez, Fernando and Fearing, Ronald},
	month = oct,
	year = {2009},
	note = {ISSN: 2153-0866},
	pages = {5027--5032},
}

@inproceedings{eguiluz_towards_2019,
	title = {Towards flapping wing robot visual perception: {Opportunities} and challenges},
	shorttitle = {Towards flapping wing robot visual perception},
	doi = {10.1109/REDUAS47371.2019.8999674},
	booktitle = {2019 {Workshop} on {Research}, {Education} and {Development} of {Unmanned} {Aerial} {Systems} ({RED} {UAS})},
	author = {Eguíluz, A. Gómez and Rodríguez-Gómez, J.P. and Paneque, J.L. and Grau, P. and de Dios, J.R. Martínez and Ollero, A.},
	month = nov,
	year = {2019},
	pages = {335--343},
}

@article{rayner_aerodynamics_1991,
	title = {On the aerodynamics of animal flight in ground effect},
	volume = {334},
	url = {https://royalsocietypublishing.org/doi/10.1098/rstb.1991.0101},
	doi = {10.1098/rstb.1991.0101},
	number = {1269},
	urldate = {2022-07-16},
	journal = {Philosophical Transactions of the Royal Society of London. Series B: Biological Sciences},
	author = {Rayner, Jeremy M. V. and Bone, Quentin},
	month = oct,
	year = {1991},
	note = {Publisher: Royal Society},
	pages = {119--128},
}

@inproceedings{zhang_design_2017,
	title = {Design optimization and system integration of robotic hummingbird},
	doi = {10.1109/ICRA.2017.7989639},
	booktitle = {2017 {IEEE} {International} {Conference} on {Robotics} and {Automation} ({ICRA})},
	author = {Zhang, Jian and Fei, Fan and Tu, Zhan and Deng, Xinyan},
	month = may,
	year = {2017},
	pages = {5422--5428},
}

@article{chukewad_robofly_2021,
	title = {{RoboFly}: {An} {Insect}-{Sized} {Robot} {With} {Simplified} {Fabrication} {That} {Is} {Capable} of {Flight}, {Ground}, and {Water} {Surface} {Locomotion}},
	volume = {37},
	issn = {1941-0468},
	shorttitle = {{RoboFly}},
	doi = {10.1109/TRO.2021.3075374},
	number = {6},
	journal = {IEEE Transactions on Robotics},
	author = {Chukewad, Yogesh M. and James, Johannes and Singh, Avinash and Fuller, Sawyer},
	month = dec,
	year = {2021},
	note = {Conference Name: IEEE Transactions on Robotics},
	pages = {2025--2040},
}

@article{ma_controlled_2013,
	title = {Controlled {Flight} of a {Biologically} {Inspired}, {Insect}-{Scale} {Robot}},
	volume = {340},
	url = {https://www.science.org/doi/full/10.1126/science.1231806},
	doi = {10.1126/science.1231806},
	number = {6132},
	urldate = {2022-05-23},
	journal = {Science},
	author = {Ma, Kevin Y. and Chirarattananon, Pakpong and Fuller, Sawyer B. and Wood, Robert J.},
	month = may,
	year = {2013},
	note = {Publisher: American Association for the Advancement of Science},
	pages = {603--607},
}

@inproceedings{tu_acting_2019,
	title = {Acting {Is} {Seeing}: {Navigating} {Tight} {Space} {Using} {Flapping} {Wings}},
	shorttitle = {Acting {Is} {Seeing}},
	doi = {10.1109/ICRA.2019.8794084},
	booktitle = {2019 {International} {Conference} on {Robotics} and {Automation} ({ICRA})},
	author = {Tu, Zhan and Fei, Fan and Zhang, Jian and Deng, Xinyan},
	month = may,
	year = {2019},
	note = {ISSN: 2577-087X},
	pages = {95--101},
}

@article{mcguire_efficient_2017,
	title = {Efficient {Optical} {Flow} and {Stereo} {Vision} for {Velocity} {Estimation} and {Obstacle} {Avoidance} on an {Autonomous} {Pocket} {Drone}},
	volume = {2},
	issn = {2377-3766},
	doi = {10.1109/LRA.2017.2658940},
	number = {2},
	journal = {IEEE Robotics and Automation Letters},
	author = {McGuire, Kimberly and de Croon, Guido and De Wagter, Christophe and Tuyls, Karl and Kappen, Hilbert},
	month = apr,
	year = {2017},
	note = {Conference Name: IEEE Robotics and Automation Letters},
	pages = {1070--1076},
}

@inproceedings{de_wagter_autonomous_2014,
	title = {Autonomous flight of a 20-gram {Flapping} {Wing} {MAV} with a 4-gram onboard stereo vision system},
	doi = {10.1109/ICRA.2014.6907589},
	booktitle = {2014 {IEEE} {International} {Conference} on {Robotics} and {Automation} ({ICRA})},
	author = {De Wagter, C. and Tijmons, S. and Remes, B. D. W. and de Croon, G. C. H. E.},
	month = may,
	year = {2014},
	note = {ISSN: 1050-4729},
	pages = {4982--4987},
}

@inproceedings{delmerico_benchmark_2018,
	title = {A {Benchmark} {Comparison} of {Monocular} {Visual}-{Inertial} {Odometry} {Algorithms} for {Flying} {Robots}},
	doi = {10.1109/ICRA.2018.8460664},
	booktitle = {2018 {IEEE} {International} {Conference} on {Robotics} and {Automation} ({ICRA})},
	author = {Delmerico, Jeffrey and Scaramuzza, Davide},
	month = may,
	year = {2018},
	note = {ISSN: 2577-087X},
	pages = {2502--2509},
}


\printindex

\end{document}